\documentclass[10pt,journal,compsoc]{IEEEtran}
\ifCLASSOPTIONcompsoc
  \usepackage[nocompress]{cite}
\else
  \usepackage{cite}
\fi
\ifCLASSINFOpdf
\else
\fi

\usepackage{subfigure,multirow,array,graphicx,amssymb,amsmath,diagbox,booktabs,xcolor}

\usepackage{ragged2e}

\newcommand{\tabincell}[2]{\begin{tabular}{@{}#1@{}}#2\end{tabular}}

\newcommand{\highlight}[1]{\textcolor{black}{#1}}
\newcommand{\Highlight}[1]{\textcolor{black}{#1}}

\begin{document}
%
\title{MOL: Joint Estimation of Micro-Expression, Optical Flow, and Landmark via Transformer-Graph-Style Convolution}
%
%
%
%

\author{Zhiwen~Shao,
        Yifan~Cheng,
        Feiran~Li,
        Yong~Zhou,
        Xuequan~Lu,
        Yuan~Xie,
        and~Lizhuang~Ma
\IEEEcompsocitemizethanks{
\IEEEcompsocthanksitem Z. Shao, Y. Cheng, F. Li, and Y. Zhou are with the School of Computer Science and Technology, China University of Mining and Technology, Xuzhou 221116, China, and also with the Mine Digitization Engineering Research Center of the Ministry of Education, Xuzhou 221116, China. Z. Shao is also with the Department of Computer Science and Engineering, The Hong Kong University of Science and Technology, Clear Water Bay, Kowloon 999077, Hong Kong, and also with the School of Computer Science, Shanghai Jiao Tong University, Shanghai 200240, China. E-mail: \{zhiwen\_shao; yifan\_cheng; feiran\_li; yzhou\}@cumt.edu.cn.
\IEEEcompsocthanksitem X. Lu is with the Department of Computer Science and Software Engineering, The University of Western Australia, Crawley WA 6009, Australia. E-mail: bruce.lu@uwa.edu.au.
\IEEEcompsocthanksitem Y. Xie is with the School of Computer Science and Technology, East China Normal University, Shanghai 200062, China. E-mail: yxie@cs.ecnu.edu.cn.
\IEEEcompsocthanksitem L. Ma is with the School of Computer Science, Shanghai Jiao Tong University, Shanghai 200240, China, also with the MoE Key Lab of Artificial Intelligence, Shanghai Jiao Tong University, Shanghai 200240, China, and also with the School of Computer Science and Technology, East China Normal University, Shanghai 200062, China. E-mail: ma-lz@cs.sjtu.edu.cn.
}
\thanks{Manuscript received April, 2023. (Corresponding authors: Yifan~Cheng, Yong~Zhou, and Lizhuang Ma.)
}}

%
%

\markboth{IEEE Transactions on Pattern Analysis and Machine Intelligence,~Vol.~X,~NO.~X,~X}%
{Shell \MakeLowercase{\textit{et al.}}: Bare Demo of IEEEtran.cls for Computer Society Journals}
%



\IEEEtitleabstractindextext{%
\begin{abstract}
\justifying Facial micro-expression recognition (MER) is a challenging problem, due to transient and subtle micro-expression (ME) actions. 
Most existing methods depend on hand-crafted features, key frames like onset, apex, and offset frames, or deep networks limited by small-scale and low-diversity datasets.
In this paper, we propose an end-to-end 
\highlight{micro-action-aware}
deep learning framework with advantages from transformer, graph convolution, and vanilla convolution. In particular, we propose a novel F5C block composed of fully-connected convolution and channel correspondence convolution to directly extract local-global features from a sequence of raw frames, without the prior knowledge of key frames. The transformer-style fully-connected convolution is proposed to extract local features while maintaining global receptive fields, and the graph-style channel correspondence convolution is introduced to model the correlations among feature patterns. Moreover, MER, optical flow estimation, and facial landmark detection are jointly trained by sharing the local-global features. The two latter tasks contribute to capturing facial subtle action information for MER, which can alleviate the impact of insufficient training data.
Extensive experiments demonstrate that our framework (i) outperforms the state-of-the-art MER methods on CASME II, SAMM, and SMIC benchmarks, (ii) works well for optical flow estimation and facial landmark detection, and (iii) can capture facial subtle muscle actions in local regions associated with MEs. \highlight{The code is available at \textit{https://github.com/CYF-cuber/MOL}.}
\end{abstract}

\begin{IEEEkeywords}
\highlight{Micro-action-aware}, transformer-style fully-connected convolution, graph-style channel correspondence convolution, micro-expression recognition, optical flow estimation, facial landmark detection
\end{IEEEkeywords}}

\maketitle

\IEEEdisplaynontitleabstractindextext

%
\IEEEpeerreviewmaketitle

\section{Introduction}

Facial micro-expression recognition (MER) is a popular task in the fields of computer vision and affective computing\highlight{~\cite{Ben2022video}}. It has applications in wide areas such as medicine, education, and criminal investigation. Micro-expressions (MEs) are subtle and involuntary that convey genuine emotions~\cite{ekman2009telling}, and contribute to the recognition of mental condition or deception of humans. Different from macro-expressions~\cite{chen2022cross,shao2024facial}, MEs are fine-grained and last only for a very short interval of time, i.e. not more than 500 milliseconds~\cite{yan2013fast}. 
In literature, MER remains a challenging problem due to the short duration, subtlety, and small-scale and low-diversity datasets of MEs. 


\begin{figure}
\centering\includegraphics[width=\linewidth]{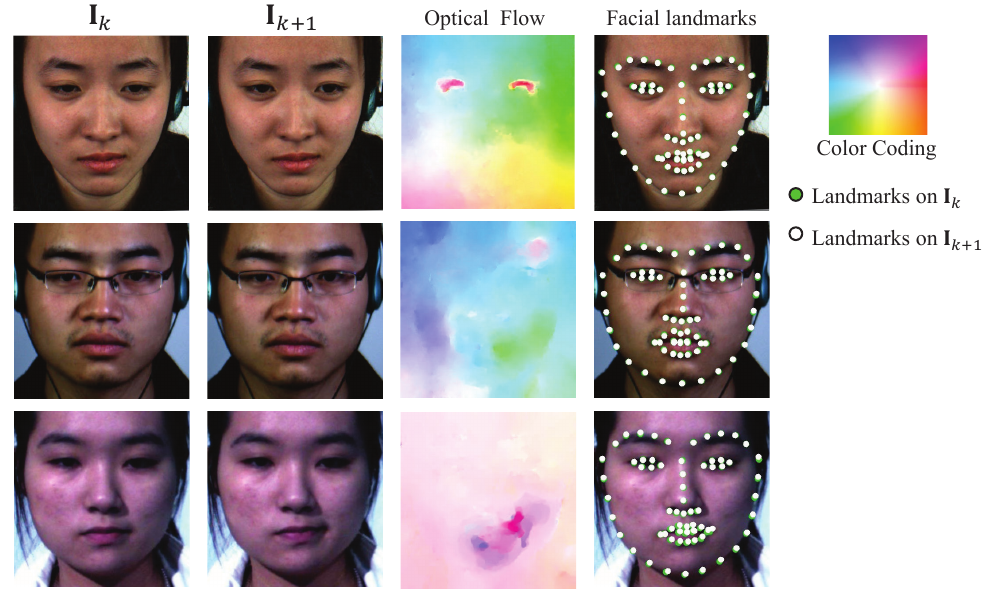}
\caption{Illustration of optical flow and facial landmark differences between two consecutive frames $\mathbf{I}_{k}$ and $\mathbf{I}_{k+1}$. We use a color coding to visualize the optical flow, in which the color of each point in the color coding denotes its displacement including orientation and magnitude to the central point. \textit{Although facial subtle muscle actions from $\mathbf{I}_{k}$ to $\mathbf{I}_{k+1}$ are hard to perceive by human eyes, they are reflected in optical flow and facial landmark differences.}
}
\label{fig:motivation}
\end{figure}

One typical way is to extract hand-crafted features containing correlated ME information. Typical hand-crafted features include optical flow and histogram of oriented optical flow (HOOF)~\cite{chaudhry2009histograms} with motion pattern, local binary patterns from three orthogonal planes (LBP-TOP)~\cite{zhao2007dynamic} with spatio-temporal information, and histogram of oriented gradients (HOG)~\cite{davison_micro-facial_2015} and histogram of image gradient orientation (HIGO)~\cite{li2018towards} with local contrast information. However, these features have limited robustness on challenging MEs with short-duration and inconspicuous motions. Besides, key frames like onset, apex, and offset frames of MEs are sometimes required in feature extraction~\cite{liong2018less}. 


Another popular solution involves the use of prevailing deep neural networks.
Khor et al.~\cite{khor2018enriched} first combined the optical flow, the derivatives of the optical flow, and the raw images as input, then used a convolutional neural network (CNN) to extract the feature of each frame and used long short-term memory (LSTM) modules to learn the temporal dynamics. However, this method relies on the pre-extracted optical flow.
Reddy et al.~\cite{reddy2019spontaneous} adopted a 3D CNN to extract features from both spatial and temporal domains, in which the performance is limited by insufficient training samples.
Xia et al.~\cite{xia2020learning} employed macro-expression recognition as an auxiliary task, in which macro-expression recognition network is used to guide the fine-tuning of MER network from both label and feature space. However, fine-grained information is not explicitly emphasized in this method.



The above methods \textit{suffer from limited capacity of hand-crafted features, requirement of key frames, or fail to thoroughly exploit the feature learning ability of deep networks due to insufficient training data}. To tackle these limitations,
we propose \highlight{to integrate automatic feature learning from raw frame sequence, capturing of facial motion information, and localization of facial fine-grained characteristics into an end-to-end framework. Considering the prevailing multi-task learning technique is convenient to guide and assist the training of main task, we design} a novel \highlight{micro-action-aware} deep learning 
framework called MOL that jointly models MER, optical flow estimation, and facial landmark detection via transformer-graph-style convolution. As illustrated in Fig.~\ref{fig:motivation}, the two latter tasks are beneficial for capturing facial subtle muscle actions associated with MEs, which relaxes the requirement of large-scale training data. Moreover, we propose a novel F5C block to directly extract local-global features from raw images, which is combined by our proposed fully-connected convolution and channel correspondence convolution. The transformer-style fully-connected convolution can extract local features while maintaining global receptive fields, and the graph-style channel correspondence convolution can model the correlations among feature map channels. Finally, we feed a sequence of pair features composed of the local-global features of consecutive two frames into a 3D CNN to achieve MER. The use of pair features rather than frame features contributes to preserving each sub-action clip, which can also be regarded as the sliding windows. The entire framework is end-to-end without any post-processing operation, and all the modules are optimized jointly.


The contributions of this paper are threefold:
\begin{itemize}
    \item We propose 
    \highlight{a micro-action-aware} joint learning framework of MER, optical flow estimation, and facial landmark detection, in which 
    pre-extracted features as well as prior knowledge of key frames are not required.
    To our knowledge, joint modeling \highlight{of automatic ME feature learning from raw frame sequence, facial motion information capturing, and facial fine-grained characteristic localization}
    via deep neural networks has not been done before.
    \item We propose a new local-global feature extractor named F5C composed by fully-connected convolution and channel correspondence convolution, which integrates the advantages of transformer, graph convolution, and vanilla convolution.
    \item Extensive experiments on benchmark datasets show that our method outperforms the state-of-the-art MER approaches, achieves competitive performance for both optical flow estimation and facial landmark detection, and can capture facial subtle muscle actions in local regions related to MEs.
\end{itemize}

\section{Related Work}

We review the previous works those are closely related to our method, including hand-crafted feature based MER, deep learning based MER, and MER with combination of hand-crafted feature and deep learning.


\subsection{Hand-Crafted Feature Based MER}

Earlier works propose hand-crafted features to try to capture fine-scale ME details. LBP-TOP~\cite{zhao2007dynamic} is a typical hand-crafted feature, which combines temporal information with spatial information from three orthogonal planes. Later, Ben et al.~\cite{ben2018learning} employed hot wheel patterns from three orthogonal planes (HWP-TOP) to make the most of the directional information. Besides, Wang et al.~\cite{wang2015efficient} proposed local binary patterns with six intersection points (LBP-SIP) to avoid repeated coding in LBP-TOP. 
Another widely used feature is histogram of oriented gradients (HOG)~\cite{davison_micro-facial_2015}, which computes gradients of image pixels. A histogram of image gradient orientation (HIGO)~\cite{li2018towards} feature is further proposed, which can maintain the invariance of geometric and optical transformation of images.

Optical flow describes the action pattern of each pixel from one frame to another frame, which is highly related to MEs. Happy et al.~\cite{happy2019fuzzy} improved histogram of oriented optical flow (HOOF)~\cite{chaudhry2009histograms} as FHOOF by collecting the action directions into angular bins based on the fuzzy membership function, and also extended FHOOF to be fuzzy histogram of optical flow orientations (FHOFO) by ignoring the action magnitude in computation. 
Liong et al.~\cite{liong2018less} introduced bi-weighted oriented optical flow (Bi-WOOF) to encode essential expressiveness of the apex frame in ME videos.

However, the extraction process of hand-crafted features often discards important information, in which the characteristics of subtle and diverse MEs are hard to be modeled. Besides, key frames of MEs are often required, which limits the applicability.




\begin{figure*}
\centering\includegraphics[width=\linewidth]{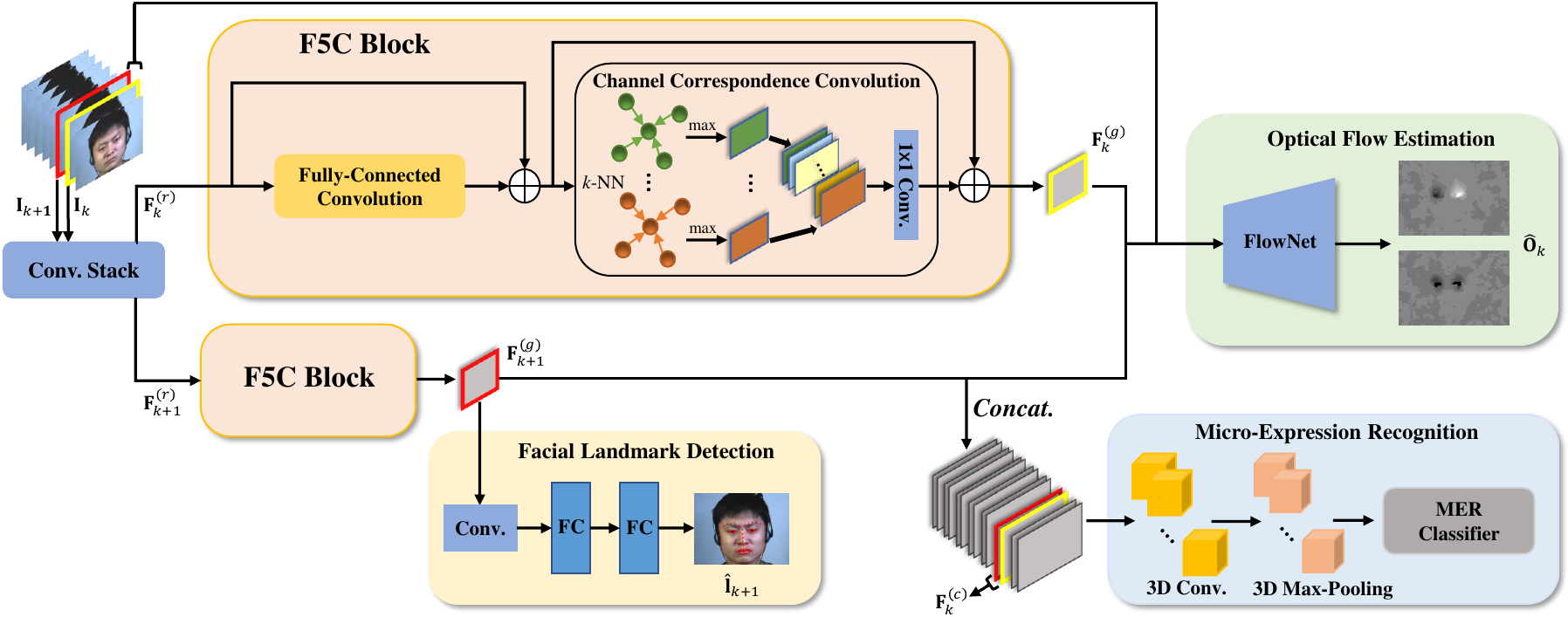}
\caption{The architecture of our MOL framework. Given a sequence of $t$ frames $\{\mathbf{I}_{0}, \mathbf{I}_{1}, \cdots, \mathbf{I}_{t-1}\}$, MOL first extracts rich feature $\mathbf{F}^{(r)}_k$ of each frame $\mathbf{I}_{k}$ by a stack of vanilla convolutional layers. For each pair of consecutive frames $\{\mathbf{I}_{k}, \mathbf{I}_{k+1}\}$,  $\mathbf{F}^{(r)}_k$ and $\mathbf{F}^{(r)}_{k+1}$ are then fed into the same F5C block to extract local-global features $\mathbf{F}^{(g)}_k$ and $\mathbf{F}^{(g)}_{k+1}$, respectively. Afterwards, $\mathbf{F}^{(g)}_{k+1}$ is fed into a facial landmark detection module to predict facial landmark locations $\hat{\mathbf{l}}_{k+1}$ of the frame $\mathbf{I}_{k+1}$, while $\mathbf{F}^{(g)}_k$, $\mathbf{F}^{(g)}_{k+1}$, $\mathbf{I}_{k}$, and $\mathbf{I}_{k+1}$ are simultaneously fed into an optical flow estimation module to predict optical flow $\hat{\mathbf{O}}_{k}$ including horizontal component and vertical component. $\mathbf{F}^{(g)}_k$ and $\mathbf{F}^{(g)}_{k+1}$ are further concatenated to be $\mathbf{F}^{(c)}_k$ as the feature of the $k$-th pair. Finally, the sequence of $t-1$ pair features $\{\mathbf{F}^{(c)}_0, \mathbf{F}^{(c)}_1, \cdots, \mathbf{F}^{(c)}_{t-2}\}$ is fed into a MER module to predict the ME category.
}
\label{fig:MOL}
\end{figure*}

\subsection{Deep Learning Based MER}

Recently, the prevailing deep learning technique has been applied to MER. 
Reddy et al.~\cite{reddy2019spontaneous} employed a 3D CNN to achieve MER, which extracts spatial and temporal information from raw image sequences. 
Lei et al.~\cite{lei2020novel} extracted shape representations based on facial landmarks, and then adopted a graph-temporal convolutional network (Graph-TCN) to capture local muscle actions of MEs. Wei et al.~\cite{wei2022novel} proposed an attention-based magnification-adaptive network (AMAN), in which a magnification attention module is used to focus on appropriate magnification levels of different MEs, and a frame attention module is used to focus on discriminative frames in a ME video.

Besides single MER task based methods, some works incorporate auxiliary tasks correlated with MER into a deep multi-task learning framework. Since action units (AUs) describe facial local muscle actions~\cite{liu2024multi,yuan2024auformer}, Xie et al.~\cite{xie2020assisted} proposed an AU-assisted graph attention convolutional network (AU-GACN), which uses graph convolutions to model the correlations among AUs so as to facilitate MER. Xia et al.~\cite{xia2020learning} used macro-expression recognition as an auxiliary task, in which macro-expression recognition network can guide the fine-tuning of MER network from both label and feature space.


Different from the above methods, we employ an end-to-end deep framework for joint learning of MER, optical flow estimation, and facial landmark detection.


\subsection{MER with Combination of Hand-Crafted Feature and Deep Learning}
Considering deep networks are limited by small-scale and low-diversity ME datasets, some approaches combine hand-crafted features with deep learning framework.
Verma et al.~\cite{verma2020learnet} proposed a dynamic image which preserves facial action information of a video, and input the dynamic image to a lateral accretive hybrid network (LEARNet). Nie et al.~\cite{nie2021geme} also generated the dynamic image of the input video, and input it to a dual-stream network with two tasks of MER and gender recognition.

Another commonly used hand-crafted feature is optical flow. Zhou et al.~\cite{zhou2019dual} calculated the optical flow between onset and apex frames of the input ME video, in which its horizontal and vertical components are fed into a dual-inception network to achieve MER. 
With the same input setting, Shao et al.~\cite{shao2023identity} achieved AU recognition and MER simultaneously, in which AU features are aggregated into ME features.
Besides, Hu et al.~\cite{hu2018multi} fused local Gabor binary pattern from three orthogonal panels (LGBP-TOP) feature and CNN feature, and then formulated MER as a multi-task classification problem, in which each category classification can be regard as a one-against-all pairwise classification problem.

All these methods require pre-extracted hand-crafted features, in which the representation power of deep networks is not thoroughly exploited.
In contrast, our network directly processes raw images, and contains a novel local-global feature extractor. Besides, instead of treating optical flow estimation as a preprocessing, we put it into a joint framework to guide the capturing of facial subtle motions.







\section{MOL for Joint Estimation of Micro-Expression, Optical Flow and Landmark}





\subsection{Overview}

Given a video clip with $t$ frames $\{\mathbf{I}_{0}, \mathbf{I}_{1}, \cdots, \mathbf{I}_{t-1}\}$, our main goal is \highlight{to design a micro-action-aware deep learning framework} to predict ME category of the overall clip, facial landmark locations $\{\hat{\mathbf{l}}_{1}, \hat{\mathbf{l}}_{2}, \cdots, \hat{\mathbf{l}}_{t-1}\}$ of the last $t-1$ frames, and optical flow $\{\hat{\mathbf{O}}_{0}, \hat{\mathbf{O}}_{1}, \cdots, \hat{\mathbf{O}}_{t-2}\}$ of the $t-1$ consecutive frame pairs $\{(\mathbf{I}_{0}, \mathbf{I}_{1}), (\mathbf{I}_{1}, \mathbf{I}_{2}), \cdots, (\mathbf{I}_{t-2}, \mathbf{I}_{t-1})\}$. We choose to directly process raw video clips without the dependence on hand-crafted features, and discard additional limitations like the prior knowledge of onset and apex frames.
Fig.~\ref{fig:MOL} illustrates the overall structure of our MOL framework.

A stack of vanilla convolutional layers are first used to extract rich feature $\mathbf{F}^{(r)}_k$ of the $k$-th frame $\mathbf{I}_{k}$ in the input video, respectively. \highlight{TABLE~\ref{tab:conv stack} shows the detailed architecture of this module.} Then, for each pair of consecutive frames $\{\mathbf{I}_{k}, \mathbf{I}_{k+1}\}$, an F5C block is used to learn local-global features $\mathbf{F}^{(g)}_k$ and $\mathbf{F}^{(g)}_{k+1}$, respectively. The local-global features are shared by three tasks for joint learning, in which optical flow estimation and facial landmark detection as auxiliary tasks are devised for promoting the main task MER in temporal and spatial domains, respectively.

\begin{table}
\centering\caption{\highlight{The structure of the stack of vanilla convolutional layers for extracting rich feature. $C_{in}$ and $C_{out}$ denote the number of input channels and output channels, respectively.}}
\label{tab:conv stack}
\begin{tabular}{c|ccccc}

\toprule
\highlight{Layer Name}& \highlight{$C_{in}$} & \highlight{$C_{out}$} &\highlight{Kernel} & \highlight{Stride} & \highlight{Padding} \\
\midrule
Convolution 1 &1 &8 &$4\times4 $&(2,2)&(0,0) \\
Convolution 2 &8 &32 &$3\times3$ &(2,2)&(0,0)\\
Convolution 3 &32 &64 &$2\times2 $&(2,2)&(1,1)\\
Convolution 4 &64 &128 &$1\times1$ &(1,1)&(0,0)\\
\bottomrule

\end{tabular}
\end{table}

To estimate the optical flow $\hat{\mathbf{O}}_{k}$ between $\mathbf{I}_{k}$ and $\mathbf{I}_{k+1}$, we simultaneously feed $\mathbf{I}_{k}$, $\mathbf{I}_{k+1}$, $\mathbf{F}^{(g)}_k$, and $\mathbf{F}^{(g)}_{k+1}$ into an optical flow estimation module. To predict the landmark locations $\hat{\mathbf{l}}_{k+1}$ of $\mathbf{I}_{k+1}$, we input $\mathbf{F}^{(g)}_{k+1}$ to a landmark detection module. Finally, we feed a sequence of $t-1$ pair features $\{\mathbf{F}^{(c)}_0, \mathbf{F}^{(c)}_1, \cdots, \mathbf{F}^{(c)}_{t-2}\}$ into a 3D CNN to predict the ME category of the whole video clip, in which $\mathbf{F}^{(c)}_k$ is the concatenation of $\mathbf{F}^{(g)}_k$ and $\mathbf{F}^{(g)}_{k+1}$. This use of pair features rather than frame features is beneficial for preserving each sub-action clip.

\subsection{F5C Block}
The architecture of our proposed F5C block is shown in the upper part of Fig.~\ref{fig:MOL}. We name this block as F5C because it consists of two main operations, fully-connected convolution (FCC) and channel correspondence convolution (CCC). 
FCC is developed from the conventional circular convolution~\cite{elliott1987handbook} by integrating the style of the prevailing transformer~\cite{vaswani_attention_2017}, which can gather local information from local receptive fields like convolutions and extract global information from the entire spatial locations like self-attention~\cite{vaswani_attention_2017}. CCC is designed to model the correlations among feature map channels in a manner of graph convolution~\cite{bruna2014spectral}. Two residual structures~\cite{he2016deep} along with FCC and CCC are beneficial for mitigating the vanishing gradient problem.  
The design of F5C integrates the merits of transformer, graph convolution, and vanilla convolution.

\begin{figure}
\centering\includegraphics[width=\linewidth]{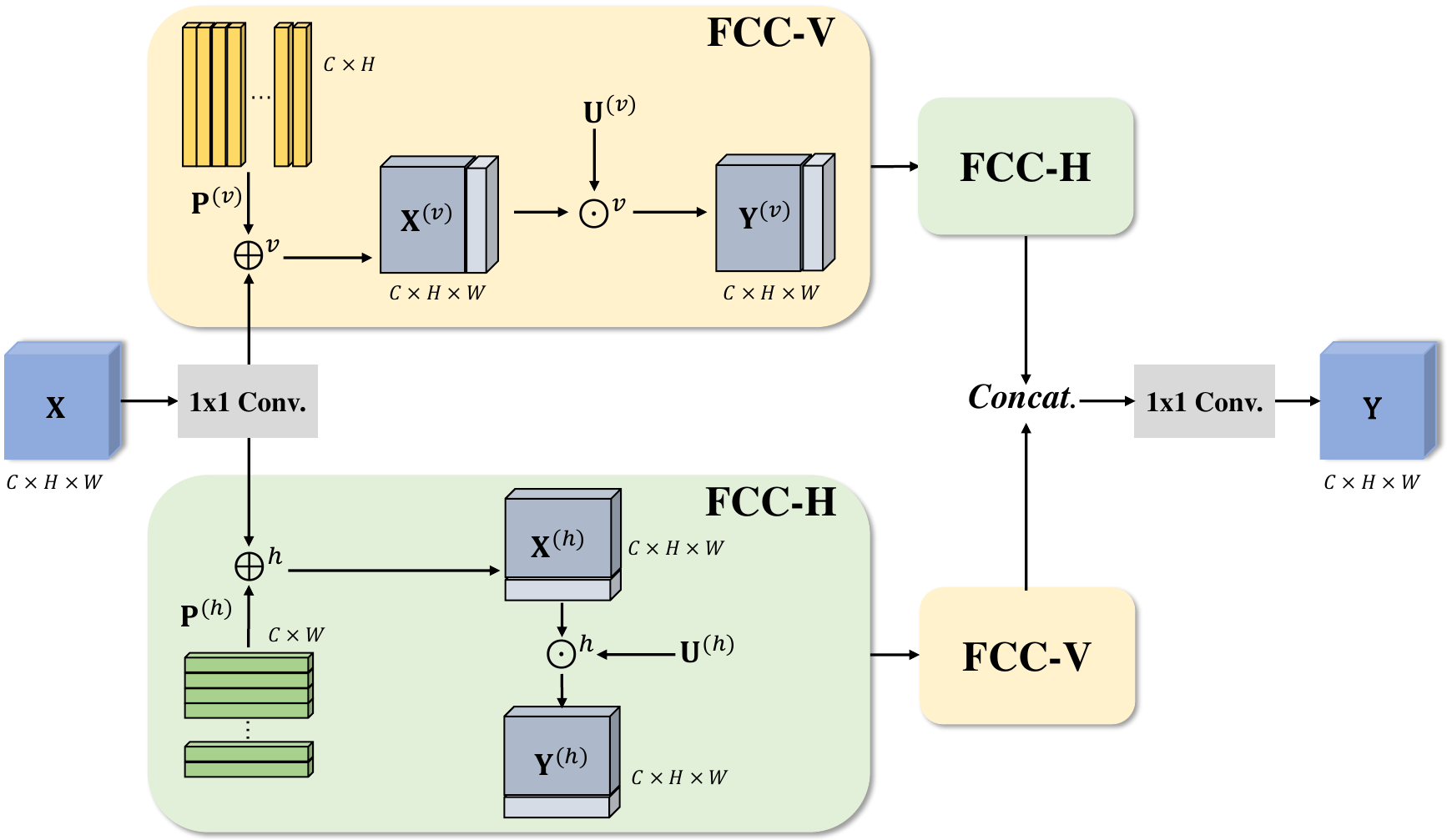}
\caption{
The structure of our proposed transformer-style fully-connected convolution. An input feature map $\mathbf{X}$ with a size of $C\times H\times W$ is first processed by vanilla $1\times 1$ convolution, and further goes through two branches, respectively, in which the first branch consists of FCC-V and FCC-H in order while the second branch uses the reverse order. Then, the outputs of the two branches are concatenated along with $1\times 1$ convolution to obtain the final output $\mathbf{Y}$ with the same size as $\mathbf{X}$.
}
\label{fig:Fully-connected}
\end{figure}

\subsubsection{Fully-Connected Convolution}


\highlight{It is known that vanilla convolution works well in extracting local features. We propose to enhance its ability of extracting global features from three aspects. First, similar to transformers~\cite{vaswani_attention_2017,dosovitskiy2021image}, we treat each column (in vertical direction) or each row (in horizontal direction) of the input as a patch, and apply positional embedding to patches to perceive contextual information. Second, we conduct circular convolution on each patch via fully-connected operation to enlarge the receptive field. Third, we perform operations in both vertical and horizontal directions to more completely cover regions. Such structure is named as transformer-style fully-connected
convolution.}




As shown in Fig.~\ref{fig:Fully-connected}, an FCC is composed of two main components, FCC-V in vertical direction and FCC-H in horizontal direction. 
It uses two branches of FCC-H after FCC-V and FCC-V after FCC-H, and then fuses two outputs by concatenation and vanilla $1\times 1$ convolution. In this way, 
the receptive field of FCC can cover positions in both vertical and horizontal directions so as to extract complete local-global features. 

Specifically, given an input $\mathbf{X}\in \mathbb{R}^{C\times H\times W}$, we conduct the $1\times 1$ convolution as a preprocessing. In FCC-V, we first employ a positional embedding~\cite{vaswani_attention_2017} to make it aware of the position information:
\begin{equation}
    \mathbf{X}^{(v)}=\mathbf{X} \oplus^v \mathbf{P}^{(v)},
\end{equation}
where $\mathbf{P}^{(v)}\in \mathbb{R}^{C\times H}$ denotes the positional embedding, and $\oplus^v$ denotes element-wise sum operation, in which $\mathbf{P}^{(v)}$ is expanded with $W$ times along horizontal direction so as to match the size of $\mathbf{X}$. 
Then, the output $\mathbf{Y}^{(v)}\in \mathbb{R}^{C\times H\times W}$ at element $(c,i,j)$ is defined as
\begin{equation}
\label{eq:Y_v}
    Y^{(v)}_{c,i,j}=
    \sum_{s=0}^{H-1} U_{c,s}^{(v)} X_{c,(i+s)\% H,j}^{(v)},
\end{equation}
where $\%$ denotes the remainder operation, and $\mathbf{U}^{(v)}\in\mathbb{R}^{C\times H}$ is a learnable parameter. The elements of $\mathbf{X}$ in vertical direction are fully-connected in a circular manner, so we name this process as fully-connected convolution-vertical (FCC-V). We represent Eq.~\eqref{eq:Y_v} as $\mathbf{Y}^{(v)}=\mathbf{U}^{(v)}\odot^{v}\mathbf{X}^{(v)}$ for simplicity.




Similarly, the process of FCC-H can be formulated as
\begin{subequations}
\begin{equation}
    \mathbf{X}^{(h)}=\mathbf{X} \oplus^h \mathbf{P}^{(h)},
\end{equation}
\begin{equation}
\label{eq:Y_h}
    Y^{(h)}_{c,i,j}=
    \sum_{s=0}^{W-1} U_{c,s}^{(h)} X_{c,i,(j+s)\% W}^{(h)},
\end{equation}
\end{subequations}
where $\mathbf{P}^{(h)}\in \mathbb{R}^{C\times W}$ is the positional embedding, $\oplus^h$ denotes the element-wise sum operation by expanding $\mathbf{P}^{(h)}$ with $H$ times along vertical direction, $\mathbf{U}^{(h)}\in\mathbb{R}^{C\times W}$ is a learnable parameter, and Eq.~\eqref{eq:Y_h} can be represented as $\mathbf{Y}^{(h)}=\mathbf{U}^{(h)}\odot^{h}\mathbf{X}^{(h)}$ for simplicity.


\subsubsection{Channel Correspondence Convolution}

Since each feature map channel encodes a type of visual pattern~\cite{zhang2018interpretable}, we propose the CCC to reason the relationships among feature map channels so as to further refine the extracted local-global features by FCC. The process of CCC is illustrated in the upper side of Fig.~\ref{fig:MOL}.

Inspired by the structure of dynamic graph convolution~\cite{wang2019dynamic}, 
we first construct a $k$-nearest neighbors ($k$-NN)~\cite{cover1967nearest} graph to find similar patterns. In particular, this directed graph is defined as $\mathcal{G} = (\mathcal{V}, \mathcal{E})$, where the vertex set \highlight{$\mathcal{V}=\{0,1,\cdots,C-1\}$} contains all the \highlight{$C$} feature map channels, and the edge set $\mathcal{E} \subseteq \mathcal{V}\times \mathcal{V}$. The size of the $i$-th feature map channel is given by $H \times W$, and we reshape it to be an $H W$-dimensional vector for the convenience of measuring similarity, denoted as $\mathbf{f}_i$. 
The neighbors of a vertex are chosen as the feature map channels with the top-$k$ cosine similarities.  

Given a directed edge $\mathbf{f}_i\leftarrow\mathbf{f}_j$, $\mathbf{f}_j$ is treated as a neighbor of $\mathbf{f}_i$. To obtain this edge feature \highlight{$\mathbf{e}_{i,j}\in \mathbb{R}^{HW}$}, we incorporate the global information encoded by $\mathbf{f}_i$ and the local neighborhood characteristics captured
by $\mathbf{f}_j-\mathbf{f}_i$:
\begin{equation}
    \label{eq:e_ijs}
    e_{i,j,s} = \mathcal{R}({\mathbf{v}_{s}^{(1)}}^{\top} \mathbf{f}_i + {\mathbf{v}_{s}^{(2)}}^{\top} ( \mathbf{f}_j-\mathbf{f}_i ) ),
\end{equation}
where $\mathcal{R}(\cdot)$ denotes the rectified linear unit (ReLU)~\cite{nair2010rectified} function, 
$\mathbf{v}_{s}^{(1)}\in \mathbb{R}^{HW}$ and $\mathbf{v}_{s}^{(2)}\in \mathbb{R}^{HW}$ are learnable parameters, $\top$ is used as the transpose of a vector, and $e_{i,j,s}$ is the $s$-th element of $\mathbf{e}_{i,j}$. Eq.~\eqref{eq:e_ijs} can be implemented by the convolution operation.

Finally, we adopt a maximum aggregation function to capture the most salient features:
\begin{equation}
\label{eq:CCC_output}
f_{i,s}^{(o)} = \max_{\{j|(i,j)\in \mathcal{E}\}}e_{i,j,s},
\end{equation}
where \highlight{$\mathbf{f}_i^{(o)}\in \mathbb{R}^{HW}$} is the output of the $i$-th feature map channel, and is further reshaped to the size of $H\times W$ and then is processed by a $1\times 1$ convolution.
With learnable parameters, our proposed CCC can adaptively model the correlations across feature map channels. \highlight{As shown in Fig.~\ref{fig:MOL} and Fig.~\ref{fig:Fully-connected}, the input and output sizes of FCC and CCC, as well as their composed F5C are all $C\times H\times W$. In this way, our proposed FCC, CCC, and F5C can all be used as plug-and-play modules.} 





\subsection{Joint Learning of Tasks}

\subsubsection{Micro-Expression Recognition}

Since MEs are subtle and short-duration, our method needs to check potential sub-action clips between each two consecutive frames so as to avoid the loss of ME clues. In this case, we concatenate local-global features $\mathbf{F}^{(g)}_k$ and $\mathbf{F}^{(g)}_{k+1}$ of each pair of consecutive frames $\{\mathbf{I}_{k}, \mathbf{I}_{k+1}\}$ to be $\mathbf{F}^{(c)}_k$, and input the sequence of $\{\mathbf{F}^{(c)}_0, \mathbf{F}^{(c)}_1, \cdots, \mathbf{F}^{(c)}_{t-2}\}$ to a 3D CNN. This feature fusion strategy can also be regarded as an application of the sliding window mechanism.

The detailed structure 
is shown in the lower right corner of Fig.~\ref{fig:MOL}. It consists of a 3D convolutional layer and a 3D max-pooling layer, and is followed by a MER classifier with two fully-connected layers.
In contrast to a 2D CNN operated in spatial domain, a 3D CNN uses 3D convolutional kernels to extract features in both spatial and temporal directions. The use of 3D max-pooling layer is to reduce the feature dimension while maintaining important information. 

Considering MER is a classification task, we employ the cross entropy loss:
\begin{equation}
\label{eq:ME_loss}
\mathcal{L}_{e}=-\sum^{n-1}_{s=0} p_s\log(\hat{p}_{s}),
\end{equation}
where $n$ is the number of ME classes, and $\hat{p}_{s}$ denotes the predicted probability that the sample is in the $s$-th class. $p_{s}$ denotes the ground-truth probability, which is $1$ if the sample is in the $s$-th class and is $0$ otherwise.

\subsubsection{Optical Flow Estimation}

Since MEs are subtle and low-intensity, it is difficult to extract related features from raw frames. Considering the optical flow contains motion information of facial muscles, which is strongly correlated to MEs, we use optical flow estimation as an auxiliary task to facilitate the learning of ME features.


\begin{figure}
\centering\includegraphics[width=\linewidth]{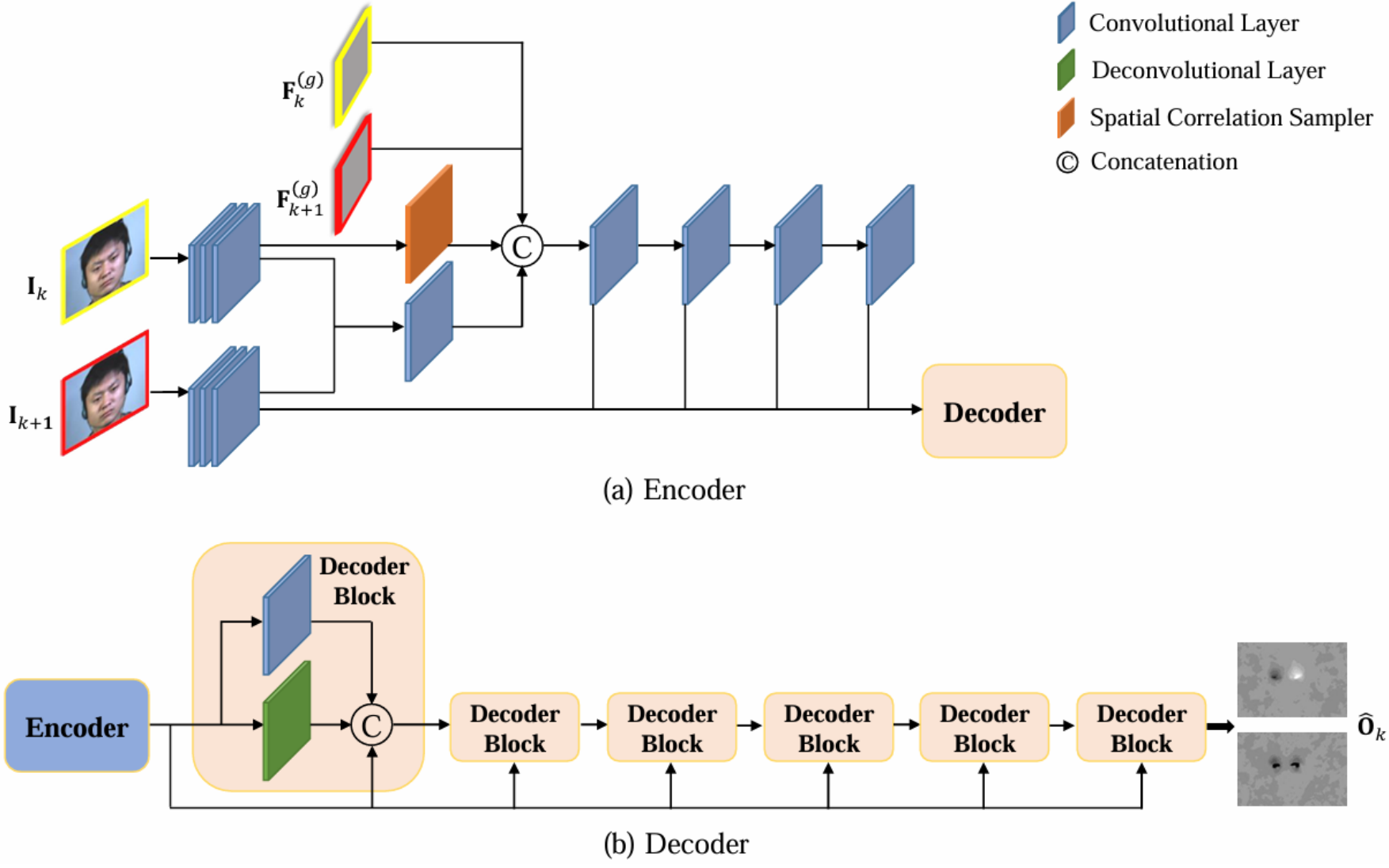}
\caption{The structure of the optical flow estimation module, which consists of (a) an encoder and (b) a decoder.}
\label{fig:Optical_Flow_Estimation}
\end{figure}

The architecture of the optical flow estimation module is detailed in Fig.~\ref{fig:Optical_Flow_Estimation}, which is based on FlowNet~\cite{dosovitskiy2015flownet} with an encoder and a decoder.
The inputs are two raw consecutive frames $\mathbf{I}_k$ and $\mathbf{I}_{k+1}$, as well as their local-global features $\mathbf{F}^{(g)}_{k}$ and $\mathbf{F}^{(g)}_{k+1}$ output by the F5C block. The encoder models the correlations between two frames and extracts multi-level features, in which the feature at each level is fed into the decoder for the final estimation of optical flow $\hat{\mathbf{O}}_k$.
The optical flow estimation loss is defined as
\begin{equation}
\label{eq:flow_loss}
\mathcal{L}_{f}=\frac{1}{t-1}\sum_{k=0}^{t-2}MSE(\mathbf{O}_k,\hat{\mathbf{O}}_k),
\end{equation}
where $\mathbf{O}_k$ denotes the ground-truth optical flow between $\mathbf{I}_k$ and $\mathbf{I}_{k+1}$, and $MSE(\cdot)$ denotes mean squared error (MSE) loss.

\subsubsection{Facial Landmark Detection}

Considering facial important regions like eyes and lips are closely related to MEs, 
we introduce another auxiliary task of facial landmark detection.
The architecture of this task module is illustrated in the bottom part of Fig.~\ref{fig:MOL}, which contains one convolutional layer and two fully-connected layers. 
The facial landmark detection loss is defined as
\begin{equation}
\label{eq:landmark_loss}
\begin{aligned}
 \mathcal{L}_{m}=\frac{1}{m(t-1) } \sum_{k=0}^{t-2}&\sum_{s=0}^{m-1} (  | l_{k+1,2s}-\hat{l}_{k+1,2s} | +\\
 &| l_{k+1,2s+1}-\hat{l}_{k+1,2s+1}  |  )/d^{(o)}_{k+1},
\end{aligned}
\end{equation}
where $\mathbf{l}_{k+1}=(l_{k+1,0},l_{k+1,1},\cdots,l_{k+1,2m-2},l_{k+1,2m-1})$ denotes the ground-truth locations of $m$ landmarks in the frame $\mathbf{I}_{k+1}$, and $l_{k+1,2s}$ and $ l_{k+1,2s+1}$ are the ground-truth x-coordinate and y-coordinate of the $s$-th landmark. Due to the differences of face sizes across samples, we use the ground-truth inter-ocular distance $d^{(o)}_{k+1}$ for normalization~\cite{shao2020deep,shao2021jaa}.


\subsubsection{\highlight{Full Loss}}

In our \highlight{micro-action-aware} joint learning framework, the full loss is composed of $\mathcal{L}_{e}$, $\mathcal{L}_{f}$, and $\mathcal{L}_{m}$:
\begin{equation}
\label{eq:total_loss}
\mathcal{L}=\mathcal{L}_{e} + \lambda_{f}  \mathcal{L}_{f}+\lambda _{m}\mathcal{L}_{m},
\end{equation}
where $\lambda_{f}$ and $\lambda_{m}$ are parameters to control the importance of optical flow estimation and facial landmark detection tasks, respectively. Besides the contributions to MER, the two auxiliary tasks can alleviate negative impact of insufficient training data.

\section{Experiments}


\subsection{Datasets and Settings}

\subsubsection{Datasets}

There are three widely used ME datasets: CASME II~\cite{yan2014casme}, SAMM~\cite{davison2016samm}, and SMIC~\cite{li2013spontaneous}.
\begin{itemize}
    \item \textbf{CASME II} contains $255$ ME videos captured from $26$ subjects, in which each video has a $280\times 340$ frame size at $200$ frames per second (FPS). These videos are selected from nearly $3,000$ elicited facial movements. Similar to the previous methods~\cite{lei2020novel,xie2020assisted}, we use ME categories of happiness, disgust, repression, surprise, and others for five-classes evaluation, and use ME categories of positive, negative, and surprise for three-classes evaluation.
    \item \textbf{SAMM} consists of $159$ ME videos from $29$ subjects, which are collected by a gray-scale camera at $200$ FPS in controlled lighting conditions without flickering. Following the previous works~\cite{lei2020novel,xie2020assisted}, we select ME categories of happiness, anger, contempt, surprise, and others for five-classes evaluation, and select ME categories of positive, negative, and surprise for three-classes evaluation.
    \item \textbf{SMIC} includes $164$ ME videos from $16$ subjects. Each video is recorded at the speed of $100$ FPS and is labeled with three ME classes (positive, negative, and surprise). It is only adopted for three-classes evaluation.
\end{itemize}
Since facial landmarks and optical flow are not annotated in these datasets, we use a powerful landmark detection library Dlib~\cite{dlib09,kazemi2014one} to detect $68$ landmarks of each frame, and use a popular optical flow algorithm TV-L1~\cite{zach2007duality} to compute optical flow between frames, both as the ground-truth annotations.


\begin{table}
\centering\caption{The number of videos for each ME class in CASME II~\cite{yan2014casme} and SAMM~\cite{davison2016samm} datasets, in which ``-'' denotes the dataset does not contain
this class, and the classes used in five-classes evaluation are highlighted with its number in bold.}
\label{tab:5-classes distribution}
\setlength\tabcolsep{16pt}
\begin{tabular}{c|*{2}{c}}
\toprule
\diagbox{Class}{Dataset} & CASME II & SAMM\\
\midrule
Happiness &\textbf{32} &\textbf{26}\\
Anger &- &\textbf{57}\\
Contempt &- &\textbf{12}\\
Disgust &\textbf{63} &9\\
Fear &2 &8\\
Repression &\textbf{27} &-\\
Surprise &\textbf{28} &\textbf{15}\\
Sadness &4 &6\\
Others &\textbf{99} &\textbf{26}\\
\bottomrule
\end{tabular}
\end{table}

\begin{table}
\centering\caption{The number of videos for each of three ME classes used in the composite dataset evaluation task. ``Composite'' denotes the combination of SMIC~\cite{li2013spontaneous}, CASME II~\cite{yan2014casme}, and SAMM~\cite{davison2016samm} datasets.}
\label{tab:3-classes distribution}
\setlength\tabcolsep{5pt}
\begin{tabular}{c|*{ 5 }{c}}
\toprule
\diagbox{Label}{Dataset} & CASME II & SAMM & SMIC & Composite\\
\midrule
Positive &32 &26 &51 &109\\
Negative &88 &92 &70 &250\\
Surprise &25 &15 &43 &83\\
\midrule
Total &145 &133 &164 &442\\
\bottomrule
\end{tabular}
\end{table}

\begin{table*}
\centering\caption{Comparison with state-of-the-art methods on CASME II~\cite{yan2014casme} and SAMM~\cite{davison2016samm}. ``DL'' denotes deep learning based methods, and ``NDL'' denotes non-deep learning based methods. ``PF'' denotes the use of pre-extracted hand-crafted features, ``RI'' denotes the use of raw images, and ``KF'' denotes the requirement on key frames such as onset, apex, and offset frames of MEs. ``Cate.'' denotes the number of ME categories. ``-'' denotes the result is not reported in its paper. The best results are highlighted in bold, and the second best results are highlighted by an underline.}
\label{tab:comparison CAMSE2 and SAMM}
\setlength\tabcolsep{10pt}
\begin{tabular}{*{3}{c}|*{3}{c}|*{3}{c}}

\toprule
\multirow{2}{*}{Method}& \multirow{2}{*}{{Paper}} & \multirow{2}{*}{{Type}} & \multicolumn{3}{c|}{CASME II} & \multicolumn{3}{c}{SAMM}\\
\cmidrule{4-9}
~&~&~&Cate. &Acc & WF1 &Cate. &Acc &WF1\\
\midrule

OFF-ApexNet~\cite{gan2019off} &SPIC'19  &DL+PF+KF &3 &88.28 &86.97 &3 &68.18 &54.23 \\

AU-GACN~\cite{xie2020assisted}&MM'20 &DL+RI &3 &71.20 &35.50 &3 &70.20 &43.30\\ 
 
GACNN~\cite{kumar2021micro} &CVPRW'21 & DL+PF &3 &\underline{89.66} &86.95 &3 &\textbf{88.72} &\underline{81.18} \\

MER-Supcon~\cite{zhi2022micro} &PRL'22 & DL+PF+KF &3 &89.65 &\underline{88.06} &3 &81.20 &71.25\\

\textbf{MOL} &Ours & DF+RI &3&\textbf{91.26} &\textbf{88.91} &3 &\underline{88.36} &\textbf{82.72} \\\midrule

SparseSampling~\cite{le2017sparsity} &TAFFC'17 & NDL &5 &49.00 &51.00 &- &- &-\\

Bi-WOOF~\cite{liong2018less} &SPIC'18 & NDL+KF &5 &58.85 &61.00 &- &- &- \\

HIGO+Mag~\cite{li2018towards} &TAFFC'18 & NDL &5 &67.21 &- &- &- &-\\
FHOFO~\cite{happy2019fuzzy} &TAFFC'19 & NDL &5 &56.64 &52.48 &-&- &-\\
DSSN~\cite{khor2019dual} &ICIP'19 & DL+PF+KF &5 &70.78 &72.97 &5 &57.35 &46.44\\

Graph-TCN~\cite{lei2020novel} & MM'20 & DL+RI+KF &5 &73.98 &72.46 &5 &\underline{75.00} &69.85 \\

MicroNet~\cite{xia2020learning} &MM'20 &DL+RI+KF  &5 &75.60 &70.10 &5 &74.10 &\textbf{73.60} \\
LGCcon~\cite{li2021joint} &TIP'21 &DL+PF+KF &5 &62.14 &60.00 &5 &35.29 &23.00\\
AU-GCN~\cite{lei2021micro} & CVPRW'21 & DL+PF+KF &5 &74.27 &70.47 &5 &74.26 &70.45 \\

GEME~\cite{nie2021geme} & Neurocomputing'21 &DL+PF &5 &75.20 &73.54 &5 &55.88 &45.38 \\

MERSiamC3D~\cite{zhao2021two} &Neurocomputing'21 &DL+PF+KF &5 &\textbf{81.89} &\textbf{83.00} &5 &68.75 &64.00 \\


MER-Supcon~\cite{zhi2022micro} &PRL'22 & DL+PF+KF &5 &73.58 &72.86 &5 &67.65 &62.51\\

AMAN~\cite{wei2022novel} &ICASSP'22 &DL+RI &5 &75.40 &71.25 &5& 68.85 &66.82 \\
SLSTT~\cite{zhang2022short} & TAFFC'22 &DL+PF &5 &75.81 &75.30 &5 &72.39 &64.00\\
Dynamic~\cite{sun2022dynamic} &TAFFC'22 &DL+RI+KF &5 &72.61 &67.00 &-&-&-\\ 
I$^2$Transformer~\cite{shao2023identity} &APIN'23 &DL+PF+KF &5 &74.26 &\underline{77.11} &5 &68.91 &\underline{73.01}\\

\textbf{MOL} &Ours &DL+RI &5 &\underline{79.23} &75.85 &5 &\textbf{76.68}&71.90 \\
\bottomrule
\end{tabular}
\end{table*}

\subsubsection{Evaluation Metrics}

For single dataset evaluation, we conduct experiments on CASME II, SAMM, and SMIC, respectively, in which the number of videos for each ME category in CASME II and SAMM are summarized in TABLE~\ref{tab:5-classes distribution}. To achieve comprehensive evaluations, we also conduct a composite dataset evaluation task~\cite{see2019megc}, in which $24$ subjects from CASME II, $28$ subjects from SAMM, and $16$ subjects from SMIC are combined into a single composite dataset with three categories used. The data distributions of 
the composite dataset evaluation task are given in TABLE~\ref{tab:3-classes distribution}. Similar to most of the previous works~\cite{lei2020novel,xie2020assisted,xia2020learning}, leave-one-subject-out (LOSO) cross-validation is employed in the single dataset evaluation and the composite dataset evaluation, in which each subject is used as the test set in turn while the remaining subjects are used as the training set. Besides, following the setting in~\cite{xie2020assisted}, we conduct a cross-dataset evaluation with three ME classes, in which CASME II and SAMM are used as the training set, respectively, and SMIC is used as the test set. 

\highlight{Following the previous works~\cite{xia2020learning,liong2019shallow},} we report accuracy (Acc) and weighted F1 score (WF1) for the single dataset evaluation and the cross-dataset evaluation, and report unweighted F1 score (UF1) and unweighted average recall (UAR) for the composite dataset evaluation.  
\highlight{WF1, UF1, and UAR are defined as
\begin{subequations}
\begin{equation}
    \label{WF1}
    WF1=\sum_{j=0}^{n-1}\frac{N_j}{N}\frac{2 TP_j}{2TP_j + FP_j + FN_j},
\end{equation}
\begin{equation}
    \label{UF1}
    UF1=\frac{1}{n} \sum_{j=0}^{n-1} \frac{2 TP_j}{2 TP_j+FP_j+FN_j},
\end{equation}
\begin{equation}
    \label{UAR}
    UAR=\frac{1}{n}\sum_{j=0}^{n-1}\frac{TP_j}{N_j},
\end{equation}
\end{subequations}
where 
$N_j$ denotes the number of samples of the $j$-th ME class, $N$ denotes the total number of samples, 
and $TP_j$, $FP_j$, and $FN_j$ denote the number of true positives, false positives, and false negatives for the $j$-th class, respectively.} 
In the following sections, all the metric results are reported in percentages, in which \% is omitted
for simplicity.


\subsubsection{Implementation Details}
In our experiments, we uniformly sample $t$ frames from a video to obtain a clip as the input of our MOL. We apply similarity transformation to each frame image based on facial landmarks, in which facial shape is preserved without changing the expression. Particularly,  
each image is aligned to $3\times144\times144$, and is randomly cropped into $3\times128\times 128$ and further horizontally flipped to enhance the diversity of training data. During testing, each image is centrally cropped into $3\times128\times 128$ to adapt to the input size.
The number of frames in the input video clip is set as $t=8$, the number of facial landmarks is set as $m=68$, \highlight{and the dimensions $C$, $H$, and $W$ of feature maps in the CCC are set as $128$, $16$, and $16$, respectively.} The trade-off parameters $\lambda_f$ and $\lambda_m$ are set to $0.1$ and $68$, respectively.  
\highlight{To set an appropriate value for the number $k$ of the nearest neighbors in the graph construction of CCC, we conduct LOSO cross-validation on the CAMSE II dataset with five classes. In each validation experiment, we select a small set from the training set as the validation set. $k$ is set as $4$ for the overall best performance on the validation sets, and is fixed for experiments on other datasets.}

Our MOL is implemented via PyTorch~\cite{paszke2019pytorch}, with the Adam optimizer~\cite{kingma2014adam}, an initial learning rate of $5\times 10^{-5}$, and a mini-batch size of $32$. Before training on ME datatsets, we pre-train MOL on a popular in-the-wild macro-expression dataset Aff-Wild2~\cite{kollias2019expression,kollias2021analysing}. It contains $323$ videos annotated by seven expression categories (neutral, anger, disgust, fear, happiness, sadness, and surprise). We also annotate the facial landmarks of each frame and the optical flow between frames by Dlib~\cite{dlib09,kazemi2014one} and TV-L1~\cite{zach2007duality}, respectively. 
Since macro-expressions are long-duration, we divide each video into multiple clips, and use each clip as the input of MOL.  
All the experiments are conducted on a single NVIDIA GeForce RTX 3090 GPU.  



\begin{table}
\centering\caption{Comparison with state-of-the-art methods on SMIC~\cite{li2013spontaneous} with three ME categories.}
\label{tab:comparison SMIC}
\setlength\tabcolsep{2.4pt}





\begin{tabular}{*{3}{c}|*{2}{c}}

\toprule
\multirow{2}{*}{Method}& \multirow{2}{*}{{Paper}} & \multirow{2}{*}{{Type}} & \multicolumn{2}{c}{SMIC}\\
\cmidrule{4-5}
~&~&~ &Acc & WF1\\
\midrule
SparseSampling~\cite{le2017sparsity} &TAFFC'17 & NDL &58.00 &60.00\\

Bi-WOOF~\cite{liong2018less} &SPIC'18 & NDL+KF &62.20 &62.00\\
HIGO+Mag~\cite{li2018towards} &TAFFC'18 & NDL &68.29 &- \\
FHOFO~\cite{happy2019fuzzy} &TAFFC'19 & NDL &51.83 &52.43\\
OFF-ApexNet~\cite{gan2019off} &SPIC'19 &DL+PF+KF &67.68 &67.09\\
MicroNet~\cite{xia2020learning}&MM'20 &DL+RI+KF &\underline{76.80} &\underline{74.40}\\
LGCcon~\cite{li2021joint} &TIP'21 &DL+PF+KF &63.41 &62.00 \\
GEME~\cite{nie2021geme} & Neurocomputing'21 &DL+PF &64.63 &61.58\\


SLSTT~\cite{zhang2022short} & TAFFC'22 &DL+PF &73.17 &72.40\\
Dynamic~\cite{sun2022dynamic} &TAFFC'22 &DL+RI+KF &76.06 &71.00 \\ 
\textbf{MOL} &Ours & DL+RI  &\textbf{80.71} &\textbf{78.81}\\
\bottomrule

\end{tabular}
\end{table}

\begin{table*}
\centering\caption{Comparison with state-of-the-art methods in terms of composite dataset evaluation~\cite{see2019megc} with three ME classes.}
\label{tab:CDE task}
\begin{tabular}{*{3}{c}|*{2}{c}|*{2}{c}|*{2}{c}|*{2}{c}}
\toprule
\multirow{2}{*}{Method}&\multirow{2}{*}{{Paper}} & \multirow{2}{*}{{Type}} & \multicolumn{2}{c|}{Composite}& \multicolumn{2}{c|}{CASME II} & \multicolumn{2}{c|}{SAMM}& \multicolumn{2}{c}{SMIC}\\
\cmidrule{4-11}
~&~&~&UF1 &UAR &UF1 &UAR &UF1 &UAR &UF1 &UAR\\
\midrule
LBP-TOP~\cite{zhao2007dynamic}&TPAMI'07 &NDL &58.82&57.85&70.26&74.29&39.54&41.02&20.00&52.80\\
Bi-WOOF~\cite{liong2018less}&SPIC'18 & NDL+KF&62.96&62.27&78.05&80.26&52.11&51.39&57.27&58.29\\
OFF-ApexNet~\cite{gan2019off}&SPIC'19 &DL+PF+KF&71.96&70.96&87.64&86.81&54.09&53.92&68.17&66.95\\
CapsuleNet~\cite{van2019capsulenet} &FG'19 &DL+RI+KF
&65.20 &65.06 &70.68 &70.18 &62.09 &59.89 &58.20 &58.77\\ 
Dual-Inception~\cite{zhou2019dual} &FG'19 &DL+PF &73.22 &72.78 &86.21 &85.60 &58.68 &56.63 &66.45&67.26  \\
STSTNet~\cite{liong2019shallow}&FG'19& DL+PF+KF&73.53&76.05&83.82&86.86&65.88&68.10&68.01&70.13\\

Part+Adversarial+EMR~\cite{liu2019neural} &FG'19 &DL+PF+KF &78.85 &78.24 &82.93 &82.09 &77.54 &71.52 &74.61 &75.30 \\

MicroNet~\cite{xia2020learning}&MM'20 &DL+RI+KF &\underline{86.40} &\textbf{85.70} &87.00&\underline{87.20} &\underline{82.50} &\underline{81.90} &\textbf{86.40}&\underline{81.00}\\

\highlight{FRL-DGT~\cite{zhai2023feature}}&\highlight{CVPR'23} &\highlight{DL+RI+KF} &\highlight{81.20} &\highlight{81.10} &\highlight{\textbf{91.90}} &\highlight{90.30} &\highlight{77.20} &\highlight{75.80} &\highlight{74.30} &\highlight{74.90} \\
\highlight{SelfME~\cite{fan2023selfme}}&\highlight{CVPR'23} &\highlight{DL+RI+KF} &\highlight{-}&\highlight{-} &\highlight{\underline{90.78}} &\highlight{\textbf{92.90}} &\highlight{-}&\highlight{-} &\highlight{80.25} &\highlight{\textbf{81.51}} \\
\textbf{MOL} &Ours &DL+RI &\textbf{87.79} &\underline{85.42} &90.08 &89.92 &\textbf{89.72} &\textbf{89.00} &\underline{81.00} &72.34\\
\bottomrule

\end{tabular}
\end{table*}


\begin{table}
\centering\caption{Comparison with state-of-the-art methods in terms of cross-dataset evaluation~\cite{xie2020assisted} with three ME classes. CASME II $\to$ SMIC denotes training on CASME II and testing on SMIC. Each method is presented with its paper in a bracket, and its results are reported by~\cite{xie2020assisted}.}
\label{tab:cross-eval}
\begin{tabular}{*{2}{c}|*{2}{c}|*{2}{c}}
\toprule
\multirow{2}{*}{Method} & \multirow{2}{*}{{Type}} & \multicolumn{2}{c|}{\tabincell{c}{CASME II\\$\to$ SMIC}}& \multicolumn{2}{c}{\tabincell{c}{SAMM\\$\to$ SMIC}} \\
\cmidrule{3-6}
~&~&Acc &WF1 &Acc &WF1\\
\midrule
\tabincell{c}{STCNN~\cite{reddy2019spontaneous}\\(IJCNN'19)}&DL+RI&31.40&19.00&32.50&19.00\\
\tabincell{c}{CapsuleNet~\cite{van2019capsulenet}\\(FG'19 )} &DL+RI+KF&32.20&15.20&32.40&17.90\\
\tabincell{c}{MER-GCN~\cite{lo2020mer}\\(MIPR'20)}&DL+RI&36.70&27.20&36.10&17.80\\
\tabincell{c}{AU-GACN~\cite{xie2020assisted}\\(MM'20)} &DL+RI &\underline{34.40}&\underline{31.90}&\textbf{45.10}&\underline{30.90}\\
\textbf{MOL} &DL+RI &\textbf{47.13} &\textbf{43.91} &\underline{44.58} &\textbf{32.32} \\
\bottomrule

\end{tabular}
\end{table}

\subsection{Comparison with State-of-the-Art Methods}
\label{ssec:comp}

We compare our MOL against state-of-the-art methods under the same evaluation setting. These methods can be divided into non-deep learning (NDL) based methods and deep learning (DL) based methods. The latter can be further classified into pre-extracted feature (PF) based methods and raw image (RI) based methods according to the type of network input. Specifically, NDL methods include LBP-TOP~\cite{zhao2007dynamic}, SparseSampling~\cite{le2017sparsity}, Bi-WOOF~\cite{liong2018less}, HIGO+Mag~\cite{li2018towards}, and FHOFO~\cite{happy2019fuzzy}. DL+PF methods include OFF-ApexNet~\cite{gan2019off}, DSSN~\cite{khor2019dual}, Dual-Inception~\cite{zhou2019dual}, STSTNet~\cite{liong2019shallow}, Part+Adversarial+EMR~\cite{liu2019neural} GACNN~\cite{kumar2021micro}, LGCcon~\cite{li2021joint}, AU-GCN~\cite{lei2021micro}, GEME~\cite{nie2021geme}, MERSiamC3D~\cite{zhao2021two}, MER-Supcon~\cite{zhi2022micro}, SLSTT~\cite{zhang2022short}, and I$^2$Transformer~\cite{shao2023identity}. DL+RI methods include STCNN~\cite{reddy2019spontaneous}, CapsuleNet~\cite{van2019capsulenet}, AU-GACN~\cite{xie2020assisted}, Graph-TCN~\cite{lei2020novel}, MER-GCN~\cite{lo2020mer}, MicroNet~\cite{xia2020learning}, AMAN~\cite{wei2022novel}, Dynamic~\cite{sun2022dynamic}\highlight{, FRL-DGT~\cite{zhai2023feature}, and SelfME~\cite{fan2023selfme}}.
Besides, some of these methods rely on key frames (KF) including onset, apex, and offset frames of MEs.

\subsubsection{Single Dataset Evaluation}

TABLE~\ref{tab:comparison CAMSE2 and SAMM} and TABLE~\ref{tab:comparison SMIC} show the comparison results on single dataset of CAMSE II, SAMM, and SMIC, respectively. It can be observed that DL based methods 
are often superior to NDL based methods, which demonstrates the strength of deep neural networks. Besides, our MOL outperforms most of the previous methods, especially for three-classes MER tasks. Note that MicroNet~\cite{xia2020learning}, GACNN~\cite{kumar2021micro}, MERSiamC3D~\cite{zhao2021two}, and I$^2$Transformer~\cite{shao2023identity} outperform MOL in a few cases. However, GACNN uses hand-crafted features, MERSiamC3D and I$^2$Transformer rely on hand-crafted features and key frames, and MicroNet requires key frames, in which their applicabilities are limited. In contrast, MOL directly processes raw frame images without requiring the prior knowledge of key frames, which is a more universal solution to MER.


\subsubsection{Composite Dataset Evaluation}

The results of composite dataset evaluation are presented in TABLE~\ref{tab:CDE task}. It can be seen that our MOL achieves competitive performance compared to state-of-the-art methods. Besides, we find that our method is the only one DL based method with raw frame images as input. 
In contrast, most previous works suffer from small-scale and low-diversity training data when using deep neural networks, in which pre-extracted hand-crafted features or key frames are required. In our method, this data scarcity issue is alleviated, due to the correlated knowledge and information provided by two auxiliary tasks of optical flow estimation and facial landmark detection.



\subsubsection{Cross-Dataset Evaluation}
We take CASME II and SAMM as the training set, respectively, in which SMIC is used as the test set. The comparison results are shown in TABLE~\ref{tab:cross-eval}. It can be seen that our MOL achieves the highest WF1 results, which demonstrates the strong generalization ability of MOL. The joint learning with optical flow estimation and facial landmark detection facilitates the extraction of ME related features, which improves the robustness \highlight{and the micro-action-aware ability} of our method for unseen samples.

\begin{table}
\centering\caption{Acc and WF1 results of MOL variants without auxiliary task modules of optical flow estimation (OFE) or facial landmark detection (FLD). These results are obtained on CASME II~\cite{yan2014casme} with five classes. The best results are highlighted in bold.}
\label{tab:effectiveness of auxiliary tasks}
\setlength\tabcolsep{22pt}
\begin{tabular}{c|*{3}{c}}
\toprule
Method & Acc & WF1\\
\midrule
MOL w/o OFE\&FLD &66.55 &58.93\\
MOL w/o FLD &76.60 &72.74\\
MOL w/o OFE &75.48 &72.31\\
\textbf{MOL} &\textbf{79.23} &\textbf{75.85}\\
\bottomrule
\end{tabular}
\end{table}

\subsection{Ablation Study}

In this section, we design ablation experiments to investigate the effectiveness of auxiliary tasks, F5C block, as well as feature fusion strategy for MER input. We conduct ablation studies on the CASME II dataset in terms of five classes.


\subsubsection{Auxiliary Tasks}
\label{sssec:ablation_auxiliary}

To investigate the effects of optical flow estimation and facial landmark detection tasks on MER, we implement MOL w/o OFE and MOL w/o FLD by removing the optical flow estimation module and the facial landmark detection module of MOL, respectively. Besides, we further implement MOL w/o OFE\&FLD by removing the both task modules. TABLE~\ref{tab:effectiveness of auxiliary tasks} shows the results of these variants of MOL. We can see that MOL w/o OFE and MOL w/o FLD both perform worse than MOL, and the performance of MOL w/o OFE\&FLD is further significantly decreased after removing both auxiliary tasks.
This is because the removal of optical flow estimation or landmark detection weakens the ability of learning facial subtle motions. We also notice that MOL w/o OFE is slightly worse than MOL w/o FLD, which indicates that optical flow estimation is more correlated with MER. In our end-to-end joint learning framework, both optical flow estimation and facial landmark detection are beneficial for MER.


\begin{table}
\centering\caption{Acc and WF1 results of MOL variants without partial or complete F5C block. The F5C block includes two main operations of fully-connected convolution (FCC) and channel correspondence convolution (CCC).}
\label{tab:effectiveness of F5C block}
\setlength\tabcolsep{23pt}
\begin{tabular}{c|*{3}{c}}
\toprule
Method & Acc & WF1\\
\midrule
MOL w/o F5C &62.90 &62.52\\
MOL w/o CCC &76.86 &74.44\\
MOL w/o FCC &65.45 &63.19\\
\textbf{MOL} &\textbf{79.23} &\textbf{75.85}\\
\bottomrule
\end{tabular}
\end{table}

\begin{table}
\centering\caption{\highlight{Acc and WF1 results of MOL variants with different number of F5C blocks in each branch of frame pair.}}
\label{tab:count of F5C block}
\setlength\tabcolsep{23pt}
\begin{tabular}{c|*{3}{c}}
\toprule
\highlight{Number of F5C} & \highlight{Acc} & \highlight{WF1}\\
\midrule
\textbf{1} &\textbf{79.23} &\textbf{75.85}\\
2 &78.68 &73.46\\
3 &78.27 &73.22\\
\bottomrule
\end{tabular}
\end{table}
\subsubsection{F5C Block}

We verify the impact of F5C block as well as its main components on MOL in TABLE~\ref{tab:effectiveness of F5C block}. When removing the whole F5C block, MOL w/o F5C only achieves the Acc of 62.90 and the WF1 of 62.52. This indicates the importance of F5C block. Furthermore, when removing FCC or CCC in the F5C block, MOL w/o FCC and MOL w/o CCC both show poor performances. It is inferred that the removal of transformer-style FCC decreases the capacity of maintaining global receptive field, and the removal of graph-style CCC may cause the failure of modeling the correlations among feature patterns. \highlight{Moreover, we implement variants of MOL using multiple stacked F5C blocks in each branch of frame pair, as presented in TABLE~\ref{tab:count of F5C block}. It can be observed that using a single FC5 block achieves the best performance. Since the training sets of ME datasets like CASME II are small-scale and low-diversity, one FC5 block is already sufficient to extract correlated ME features.}


\begin{table}
\centering\caption{
\highlight{Acc and WF1 results of MOL variants with different structures of FCC.}
}
\label{tab:FCC vs Transformer}
\setlength\tabcolsep{18pt}
\begin{tabular}{c|*{3}{c}}
\toprule
\highlight{Method} & \highlight{Acc} & \highlight{WF1} \\
\midrule
Vanilla Transformer~\cite{vaswani_attention_2017} &76.30 &73.61 \\
FCC-V &78.04 &74.58 \\
FCC-H &77.64 &74.16 \\
\textbf{FCC} &\textbf{79.23} &\textbf{75.85}\ \\

\bottomrule
\end{tabular}
\end{table}

\begin{table}
\centering\caption{Acc and WF1 results of MOL variants with different feature fusion strategies for MER input. $\mathbf{F}^{(c)}_k$ is the concatenation of $\mathbf{F}^{(g)}_k$ and $\mathbf{F}^{(g)}_{k+1}$, $\mathbf{F}^{(a)}_k$ is the element-wise addition of $\mathbf{F}^{(g)}_k$ and $\mathbf{F}^{(g)}_{k+1}$ \highlight{, and $\mathbf{F}^{(s)}_k$ is the element-wise subtraction of $\mathbf{F}^{(g)}_k$ and $\mathbf{F}^{(g)}_{k+1}$}.}
\label{tab:feature_fusion}
\setlength\tabcolsep{19pt}
\begin{tabular}{c|*{2}{c}}
\toprule
Input & Acc & WF1\\
\midrule
$\{\mathbf{F}^{(g)}_0, \mathbf{F}^{(g)}_1, \cdots, \mathbf{F}^{(g)}_{t-2}\}$ &75.04 &72.41\\
$\{\mathbf{F}^{(g)}_{1}, \mathbf{F}^{(g)}_{2}, \cdots, \mathbf{F}^{(g)}_{t-1}\}$  &73.80 &72.11 \\
$\{\mathbf{F}^{(g)}_{0}, \mathbf{F}^{(g)}_{1}, \cdots, \mathbf{F}^{(g)}_{t-1}\}$  &74.65& 71.94 \\
$\{\mathbf{F}^{(a)}_0, \mathbf{F}^{(a)}_1, \cdots, \mathbf{F}^{(a)}_{t-2}\}$ &72.51 &69.90\\
\highlight{$\{\mathbf{F}^{(s)}_0, \mathbf{F}^{(s)}_1, \cdots, \mathbf{F}^{(s)}_{t-2}\}$} &\highlight{72.13} &\highlight{69.34}\\
$\{\mathbf{F}^{(c)}_0, \mathbf{F}^{(c)}_1, \cdots, \mathbf{F}^{(c)}_{t-2}\}$  &\textbf{79.23} &\textbf{75.85} \\
\bottomrule
\end{tabular}
\end{table}

\highlight{\subsubsection{FCC vs. Transformer}
To verify the effect of transformer-style FCC, we implement variants of MOL by replacing the whole FCC block with vanilla transformer, FCC-V, and FCC-H, respectively. The results are shown in TABLE~\ref{tab:FCC vs Transformer}. It can be seen that the complete FCC structure outperforms the vanilla transformer. Besides, FCC-V or FCC-H with one-directional perception still performs better. This is due to the insufficiency of ME training data, in which the power of transformer is limited, while our proposed FCC has a stronger learning ability of both local and global features. The fully-connected convolution in both vertical and horizontal directions works the best in terms of perceiving micro-actions related to MEs.}  

\subsubsection{Feature Fusion Strategy for MER Input}

As shown in Fig.~\ref{fig:MOL}, local-global features $\mathbf{F}^{(g)}_k$ and $\mathbf{F}^{(g)}_{k+1}$ of consecutive frames $\mathbf{I}_k$ and $\mathbf{I}_{k+1}$ are concatenated to be $\mathbf{F}^{(c)}_k$ as the feature of the $k$-th frame pair, then the sequence of $t-1$ pair features $\{\mathbf{F}^{(c)}_0, \mathbf{F}^{(c)}_1, \cdots, \mathbf{F}^{(c)}_{t-2}\}$ is fed into the MER module. Here we investigate the effects of different feature fusion strategies for MER input, as shown in TABLE~\ref{tab:feature_fusion}. If we do not fuse the local-global features of each two consecutive frames, the performances are all degraded for three types of inputting the first $t-1$ frame features $\{\mathbf{F}^{(g)}_0, \mathbf{F}^{(g)}_1, \cdots, \mathbf{F}^{(g)}_{t-2}\}$, inputting the last $t-1$ frame features $\{\mathbf{F}^{(g)}_{1}, \mathbf{F}^{(g)}_{2}, \cdots, \mathbf{F}^{(g)}_{t-1}\}$, and inputting all the $t$ frame features $\{\mathbf{F}^{(g)}_{0}, \mathbf{F}^{(g)}_{1}, \cdots, \mathbf{F}^{(g)}_{t-1}\}$. This is due to the sub-action clips between each two consecutive frames, which are highly related to MEs. We also implement another \highlight{two} feature fusion strategies, element-wise addition \highlight{and element-wise subtraction of} frame features. However, \highlight{both} performances become much worse, which indicates that concatenation is a better way to preserve sub-action clips. 

\subsubsection{\Highlight{Number of Input Frames}}

\Highlight{Here we investigate the impacts of different numbers of input frames on our MOL. 
Due to the characteristic of processing pairs of consecutive frames in the input video clip, we can directly feed a video clip composed of only the onset and apex frames into MOL without changing the network structure. 
TABLE~\ref{tab:Frame num} shows the results of different inputs to MOL, including key frames only and video clips with different frame amounts, in which the latter are sampled at equal intervals from the raw videos. Compared to the results of inputting 8 frames, the performance of inputting onset and apex frames shows a slight improvement, which can be attributed to the fact that the prior key frames contain the most prominent ME motion characteristics. When inputting 4 frames, the performance is significantly lower than the cases of 8 or 16 frames. This is because when sampling at equal intervals, if the number of sampled frames is too small, the obtained video clips are likely to miss some frames with high ME intensities. When inputting 8 or 16 frames, the results are relatively close. This is because the sampled clips already contain enough ME frames with high intensities. With the strong feature capture ability of F5C block and the joint framework, our MOL is competitive to those methods relying on key frames.}

\begin{table}
\centering\caption{
\Highlight{Acc and WF1 results of our MOL with different numbers of input frames on CASME II~\cite{yan2014casme}.}
}

\label{tab:Frame num}
\setlength\tabcolsep{19.1pt}
\begin{tabular}{c|*{3}{c}}
\toprule
\Highlight{Number of Input Frames} & \Highlight{Acc} & \Highlight{WF1} \\
\midrule
2 (Onset and Apex)  &\textbf{79.88} &\textbf{76.03} \\
4 &75.17 &72.81 \\
\textbf{8} &\underline{79.23} &\underline{75.85} \\
16 &78.98 &75.25 \\

\bottomrule
\end{tabular}
\end{table}

\begin{table}
\centering\caption{Average EPE results of different optical flow estimation methods on CASME II~\cite{yan2014casme}. The best results are highlighted in bold, and the second best results are highlighted by an underline.}
\label{tab:MOL for OFE}
\setlength\tabcolsep{29pt}
\begin{tabular}{c|c}
\toprule
Method & Average EPE \\
\midrule
UnsupFlownet~\cite{jjyu2016unsupflow} &1.048 \\
RAFT~\cite{teed2020raft} & \textbf{0.465}\\
\midrule
MOL w/o MER\&FLD & 1.145\\
MOL w/o FLD & 0.972\\
MOL w/o MER & 0.839\\
\textbf{MOL} & \underline{0.650}\\
\bottomrule
\end{tabular}
\end{table}

\subsection{MOL for Optical Flow Estimation and Facial Landmark Detection}
We have validated the contributions of optical flow estimation and facial landmark detection to MER in Sec.~\ref{sssec:ablation_auxiliary}. In this section, we also investigate the effectiveness of MER for these two tasks in our \highlight{micro-action-aware} joint learning framework.

\subsubsection{MOL for Optical Flow Estimation}
We implement a baseline method MOL w/o MER\&FLD which only achieves the optical flow estimation task by removing the MER and facial landmark detection modules. Besides, we implement MOL w/o MER and MOL w/o FLD by discarding MER and facial landmark detection, respectively. We also compare with two recent deep learning based optical flow estimation methods UnsupFlownet~\cite{jjyu2016unsupflow} and RAFT~\cite{teed2020raft} with code released. Average end-point error (EPE) is reported as the evaluation metric.

TABLE~\ref{tab:MOL for OFE} shows the average EPE results on the CASME II benchmark. With the help of MER and facial landmark detection, MOL outperforms MOL w/o MER\&FLD by a large margin of 0.495. When only removing one module, the results of MOL w/o MER and MOL w/o FLD are also both better than MOL w/o MER\&FLD. It is demonstrated that MEs and facial landmarks are closely related to the motion patterns captured by optical flow. 
Furthermore, despite being designed for MER, our MOL shows competitive results compared with the state-of-the-art optical flow estimation methods.

\subsubsection{MOL for Facial Landmark Detection}

We implement MOL w/o MER\&OFE as a baseline method which only achieves the facial landmark detection task without the MER and optical flow estimation modules. Besides, we implement MOL w/o MER and MOL w/o OFE by removing MER and optical flow estimation, respectively. We also compare with two popular facial landmark detection methods TCDCN~\cite{zhang2015learning} and HRNetV2~\cite{wang2020deep} with code released. We report two metrics, inter-ocular distance normalized mean error, and failure rate, in which the mean error larger than 10\% is treated as a failure. For simplicity, \% is omitted in the following mean error and failure rate results.

TABLE~\ref{tab:MOL for FLD} shows the landmark detection results on CASME II. We can see that MOL w/o OFE and MOL w/o MER both perform better than the baseline MOL w/o MER\&OFE, which proves that MER and optical flow estimation both contribute to facial landmark detection. Moreover, MOL outperforms all the above three variants, which demonstrates that our joint framework is beneficial for improving the performance of facial landmark detection. Besides, the comparison with TCDCN and HRNetV2 indicates the superiority of our MOL for landmark detection.

\begin{table}
\centering\caption{Mean error and failure rate results of different facial landmark detection methods on CASME II~\cite{yan2014casme}.}
\label{tab:MOL for FLD}
\setlength\tabcolsep{13pt}
\begin{tabular}{c|*{2}{c}}
\toprule
Method & Mean Error & Failure Rate\\
\midrule
TCDCN~\cite{zhang2015learning} &6.75 &3.84 \\
HRNetV2~\cite{wang2020deep} &\underline{4.68} &2.99 \\
\midrule
MOL w/o MER\&OFE &5.46 &7.87\\
MOL w/o OFE &5.22 &6.60\\
MOL w/o MER &4.89 &\underline{2.42}\\
\textbf{MOL} &\textbf{2.35} &\textbf{2.13}\\
\bottomrule
\end{tabular}
\end{table}

\begin{figure*}
\centering\includegraphics[width=\linewidth]{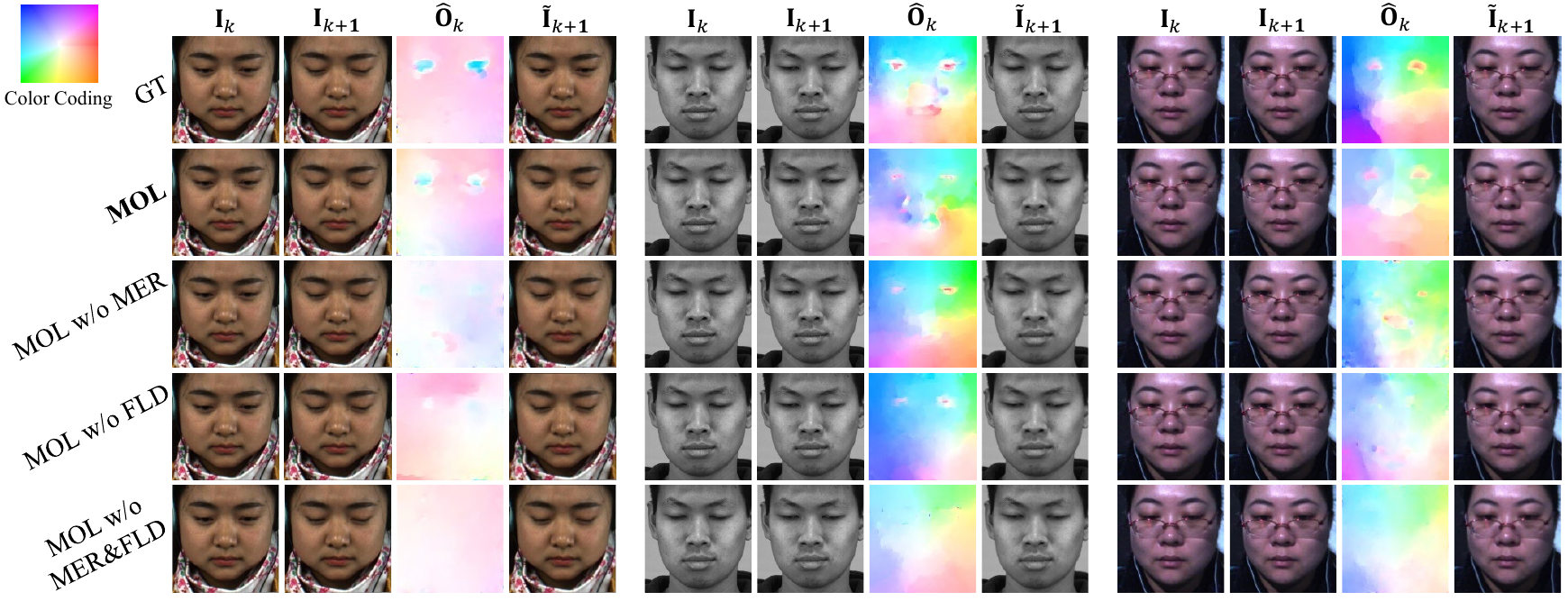}
\caption{Visualization of optical flow estimation results for three example frame pairs $\mathbf{I}_{k}$ and $\mathbf{I}_{k+1}$ from CASME II~\cite{yan2014casme}, SAMM~\cite{davison2016samm}, and SMIC~\cite{li2013spontaneous}, respectively. $\hat{\mathbf{O}}_k$ is estimated optical flow, and $\tilde{\mathbf{I}}_{k+1}$ is warped from $\mathbf{I}_{k+1}$ by $\hat{\mathbf{O}}_k$. The color coding with its central point as the original point is used to visualize the optical flow, in which the color of each point denotes its displacement including orientation and magnitude to the original point. ``GT'' denotes the ground-truth optical flow.}
\label{fig:of_viz}
\end{figure*}

\subsection{Visual Results}

To prove that our proposed method can pay attention to the subtle movements related to MEs, we visualize the estimated optical flow of different methods on several example frame pairs in Fig.~\ref{fig:of_viz}. For a better view, we use $\hat{\mathbf{O}}_k$ with horizontal component $\hat{\mathbf{A}}_k$ and vertical component $\hat{\mathbf{B}}_k$ to warp $\mathbf{I}_{k+1}$, in which the warped image $\tilde{\mathbf{I}}_{k+1}$ at each pixel position $(a,b)$ is formulated as
\begin{equation}
    \tilde{I}_{k+1,a,b}= I_{k+1,a+\hat{A}_{k,a,b},b+\hat{B}_{k,a,b}},
\end{equation}
where bilinear sampling is employed, and $\tilde{\mathbf{I}}_{k+1}$ is expected to be similar to $\mathbf{I}_{k}$. 
We can see that our MOL achieves the most accurate optical flow estimations, 
in which the slightly \Highlight{closed} eyes in the first example, the slightly shaking \Highlight{eyes, nose and mouth} in the second example, and the slightly \Highlight{open} eyes in the third example are all captured.
When the modules of MER or facial landmark detection are removed, many nonexistent motion patterns are estimated. 
Therefore, our MOL can capture facial subtle muscle movements associated with MEs due to the introduction of optical flow estimation.

\begin{figure}
\centering\includegraphics[width=\linewidth]{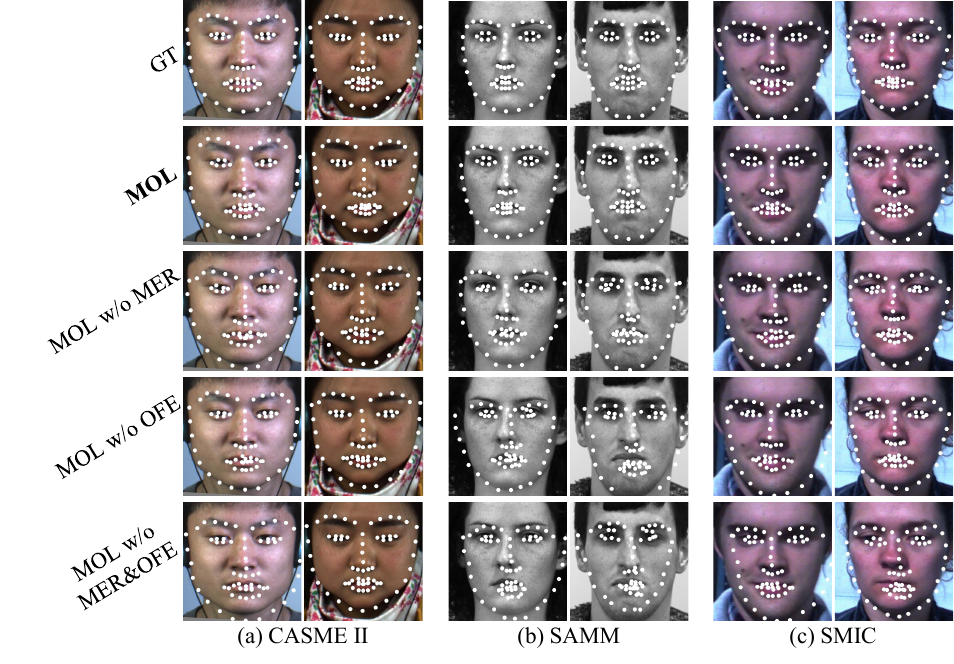}
\caption{Facial landmark detection results for example images from CASME II~\cite{yan2014casme}, SAMM~\cite{davison2016samm}, and SMIC~\cite{li2013spontaneous}. ``GT'' denotes the ground-truth locations of facial landmarks.}
\label{fig:landmark_viz}
\end{figure}

We also show facial landmark detection results on several example images in Fig.~\ref{fig:landmark_viz}. We can observe that our MOL more accurately localize facial landmarks than other variants especially for the landmarks in regions of eyes and mouth. With the help of landmark detection, our MOL can capture important facial local regions where ME actions often occur.


\section{Conclusion}

In this paper, we have developed a 
\highlight{micro-action-aware}
deep learning framework for joint MER, optical flow estimation, and facial landmark detection, in which the three tasks contribute to each other by sharing features. In addition, we have proposed the F5C block including fully-connected convolution and channel correspondence convolution to extract local-global features and model the correlations among feature channels. Our framework is end-to-end, and does not depend on pre-extracted features and key frames, which is a new solution to MER with good applicability.

We have compared our approach with state-of-the-art methods on the challenging SMIC, CASME II, and SAMM benchmarks. It is demonstrated that our approach outperforms previous MER works in terms of single dataset evaluation, composite dataset evaluation, and cross-dataset evaluation. In addition, we have conducted an ablation study which indicates that main components in our framework are all beneficial for MER. 
Besides, the experiments on optical flow estimation and facial landmark detection show competitive performance of our method, and the visual results demonstrate that our approach can capture facial subtle muscle motions in local regions associated with MEs. 

\highlight{Considering the MER task requires only a single ME in an input video, in the future work we will explore end-to-end ME spotting and recognition. In this case, multiple MEs are simultaneously localized and recognized. This is a challenging task, and is also promising for the development of the ME field.}




%

%

\ifCLASSOPTIONcompsoc
  \section*{Acknowledgments}
\else
  \section*{Acknowledgment}
\fi

This work was supported by the National Natural Science Foundation of China (No. 62472424), the China Postdoctoral Science Foundation (No. 2023M732223), and the Hong Kong Scholars Program (No. XJ2023037/HKSP23EG01). It was also partially supported by the National Natural Science Foundation of China (No. 62272461, No. 72192821, No. 62472282, No. 62222602, and No. 62176092).

\ifCLASSOPTIONcaptionsoff
  \newpage
\fi



\bibliographystyle{IEEEtran}
\bibliography{references}

%

\begin{IEEEbiography}[{\includegraphics[width=1in,height=1.25in,clip,keepaspectratio]{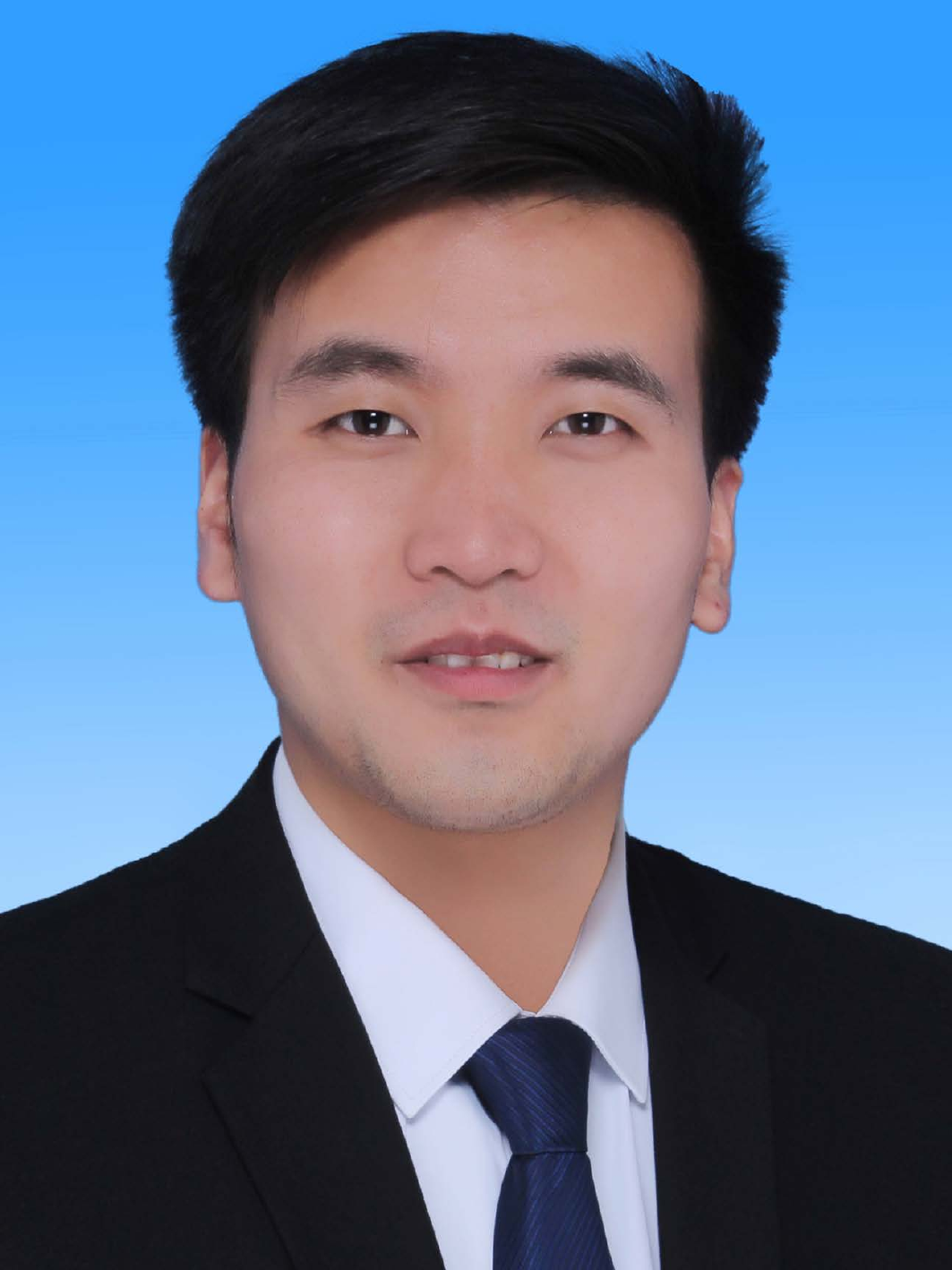}}]{Zhiwen Shao} is currently an Associate Professor with the China University of Mining and Technology, China. He received the B.Eng. degree and the Ph.D. degree in Computer Science and Technology from the Northwestern Polytechnical University, China and the Shanghai Jiao Tong University, China in 2015 and 2020, respectively. His research interests lie in computer vision and affective computing. He has served as an Area Chair for ACM MM, an Associate Editor for TVC, and a Publication Chair for CGI.
\end{IEEEbiography}

\begin{IEEEbiography}[{\includegraphics[width=1in,height=1.25in,clip,keepaspectratio]{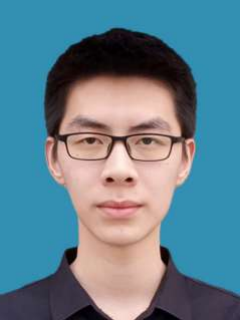}}]{Yifan Cheng} received the B.Eng. and M.S. degrees from the School of Computer Science and Technology, China University of Mining and Technology, China in 2022 and 2025, respectively. He will purse the Ph.D. degree at the School of Computer Science and Technology, East China Normal University, China. His research interest lies in facial expression recognition. 
\end{IEEEbiography}

\begin{IEEEbiography}[{\includegraphics[width=1in,height=1.25in,clip,keepaspectratio]{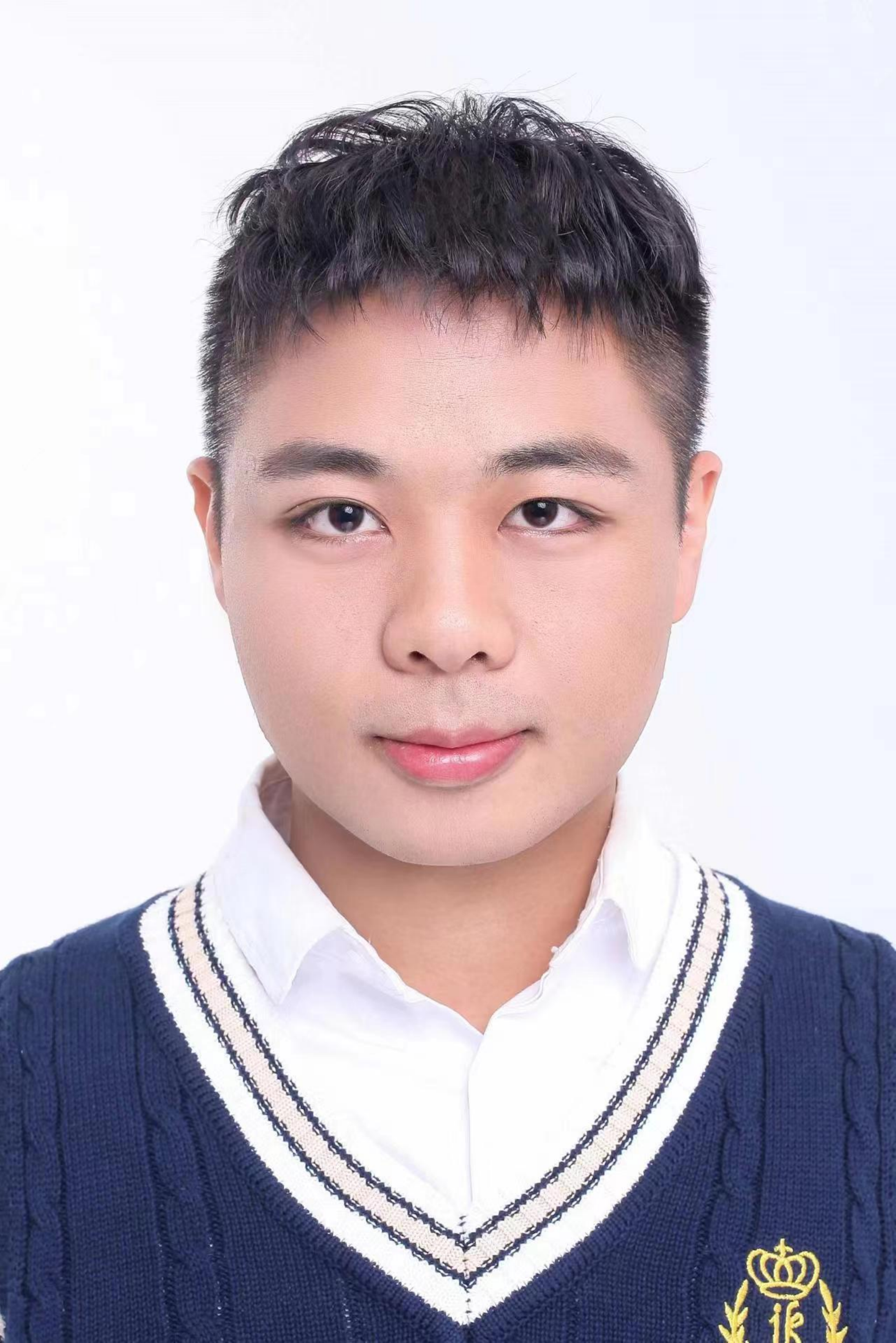}}]{Feiran Li} received the B.Eng. degree from the School of Computer Science and Technology, China University of Mining and Technology, China in 2023. He is currently pursuing the Ph.D. degree at the University of Chinese Academy of Sciences, China. His research interest lies in facial expression recognition. 
\end{IEEEbiography}

\begin{IEEEbiography}[{\includegraphics[width=1in,height=1.25in,clip,keepaspectratio]{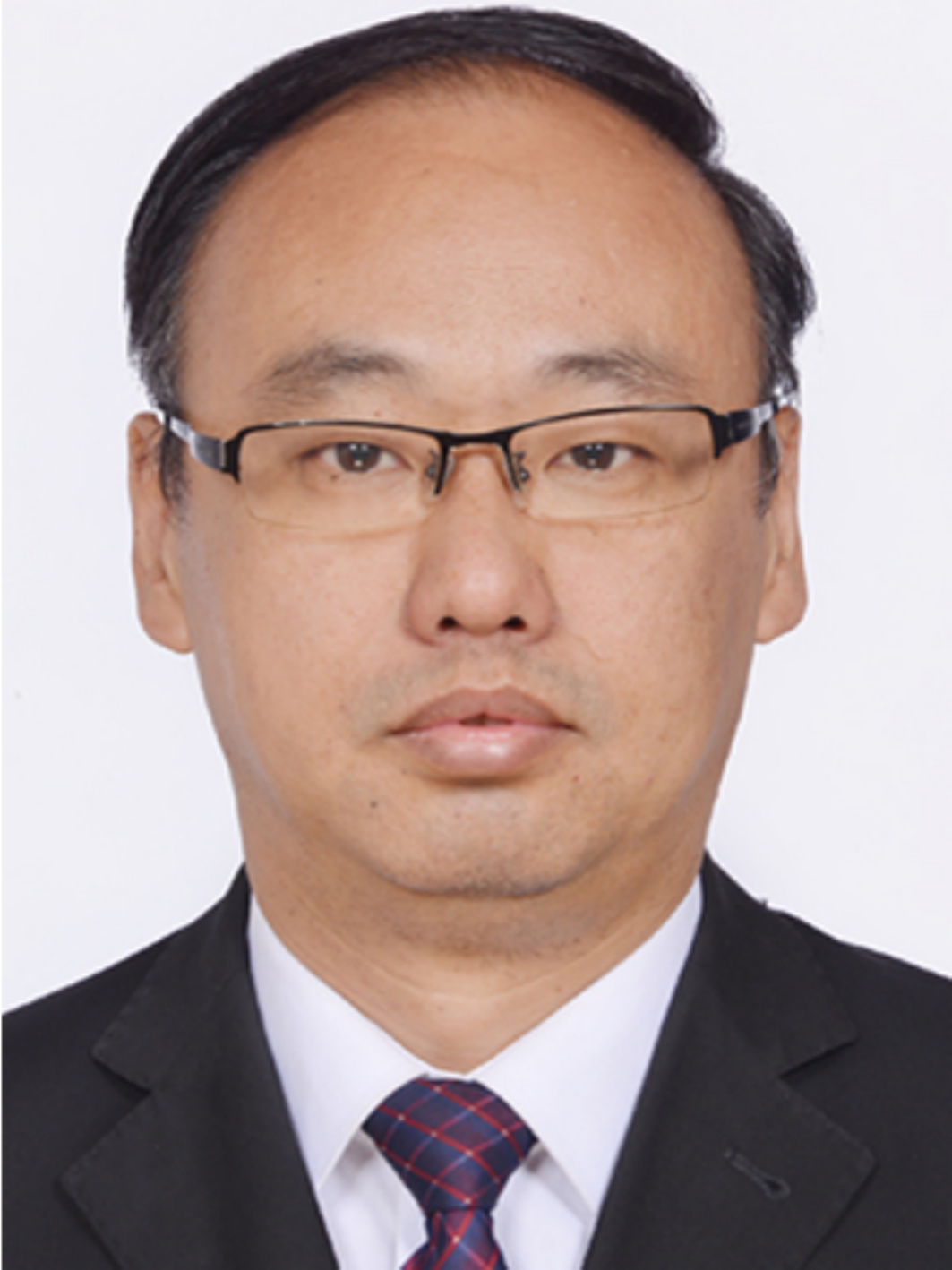}}]{Yong Zhou} is currently a Full Professor with the School of Computer Science and Technology, China University of Mining and Technology, China. He received the M.S. and Ph.D. degrees in Control Theory and Control Engineering from the China University of Mining and Technology, China in 2003 and 2006, respectively. His research interests include machine learning, intelligence optimization, and data mining. He has been serving as an Associate Editor for ACM TOMM.
\end{IEEEbiography}

\begin{IEEEbiography}[{\includegraphics[width=1in,height=1.25in,clip,keepaspectratio]{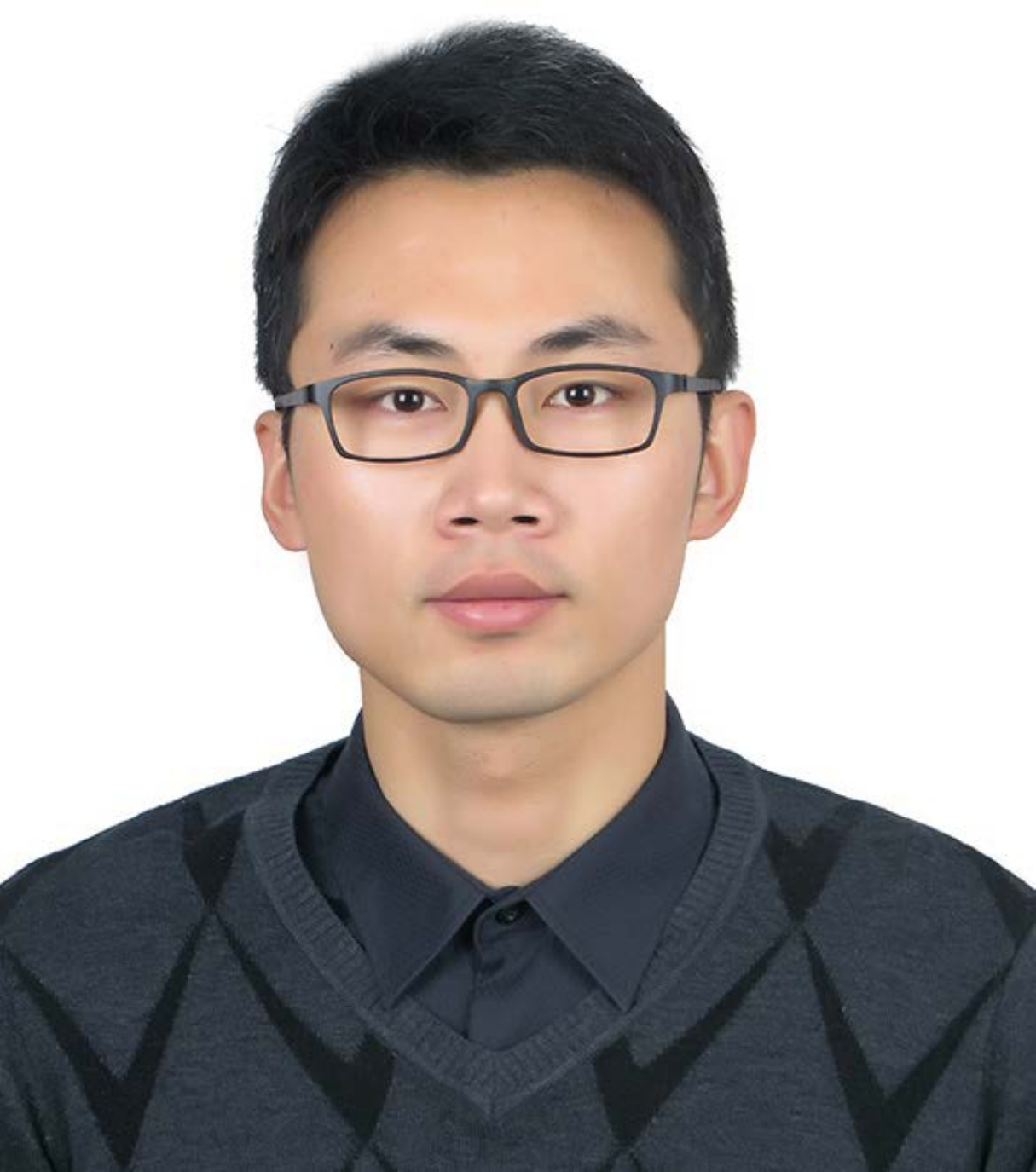}}]{Xuequan Lu}
 is currently a Senior Lecturer with the Department of Computer Science and Software Engineering, The University of Western Australia, Australia. He spent more than two years as a Research Fellow in Singapore. His research interests mainly fall into visual computing. He has served as an Associate Editor for Neurocomputing and TVC, a Program Co-Chair for ICVR 2023, as well as a Session/Area Chair for ICME 2023, ICONIP 2022, and ICONIP 2020.
\end{IEEEbiography}

\begin{IEEEbiography}[{\includegraphics[width=1in,height=1.25in,clip,keepaspectratio]{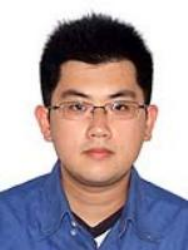}}]{Yuan Xie} is currently a Full Professor with the School of Computer Science and Technology, East China Normal University, China. He is the recipient of the National Science Fund for Excellent Young Scholars. He received the Ph.D. degree in Pattern Recognition and Intelligent Systems from the Institute of Automation, Chinese Academy of Sciences in 2013. His research interests include computer vision, machine learning, and pattern recognition. He has been serving as a SPC member for IJCAI and CIKM.
\end{IEEEbiography}

\begin{IEEEbiography}[{\includegraphics[width=1in,height=1.25in,clip,keepaspectratio]{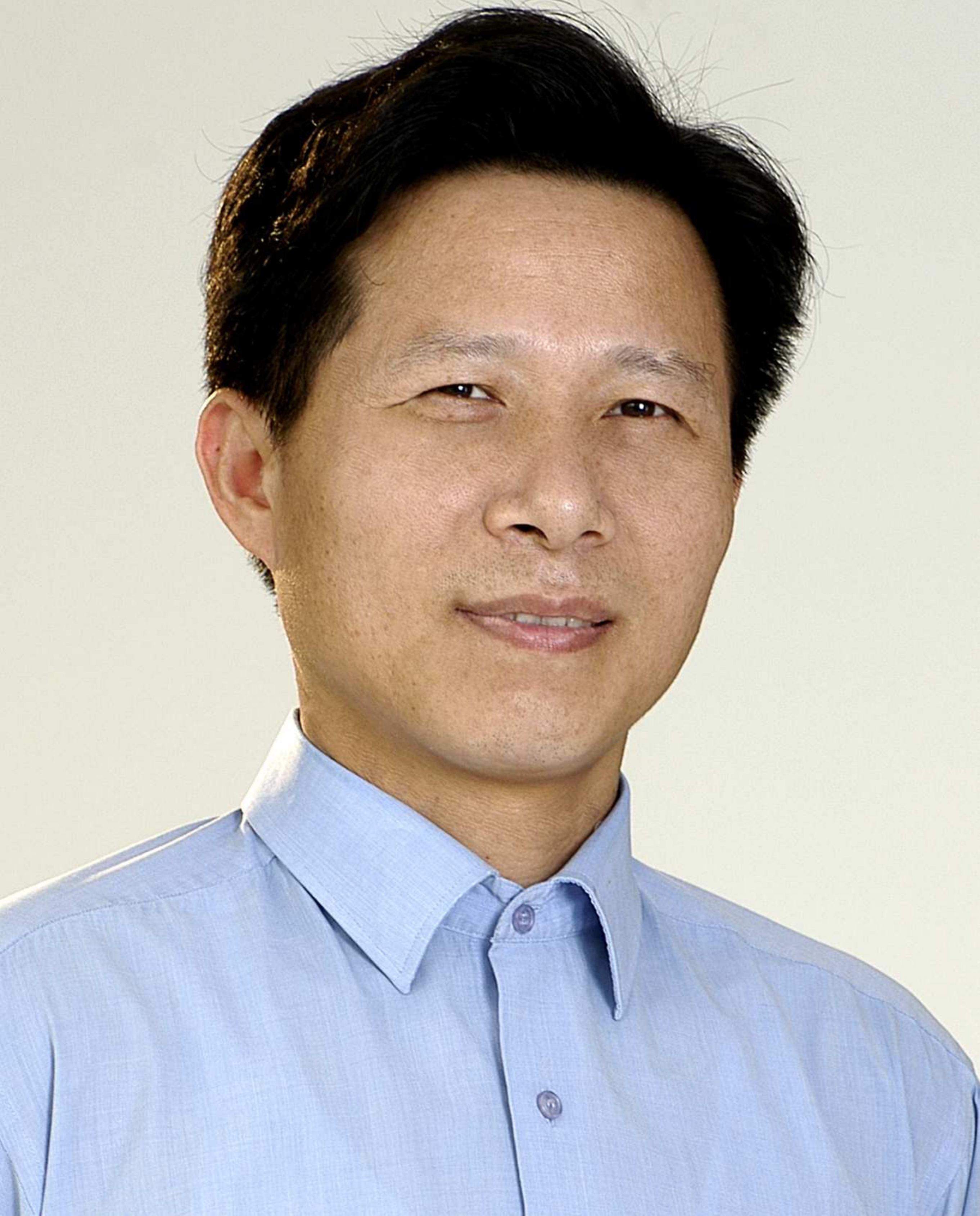}}]{Lizhuang Ma} is currently a Distinguished Professor with the Department of Computer Science and Engineering, Shanghai Jiao Tong University, China. He is the recipient of the National Science Fund for Distinguished Young Scholars. He received the B.S. and Ph.D. degrees from the Zhejiang University, China in 1985 and 1991, respectively. His research interests include computer graphics, computer animation, and theory and applications for computer graphics, CAD/CAM.
\end{IEEEbiography}


%
%
%




\end{document}